\documentclass[sigconf,nonacm]{acmart}

\usepackage{graphicx}
\usepackage{tikz}
\usepackage{cleveref}
\usepackage{xspace}
\usepackage{subcaption}
\usepackage{hyperref}
\usepackage{color}
\usepackage{listings}
\usepackage[ruled,vlined]{algorithm2e}
\usepackage{overpic}
\usepackage{amsmath}
\usepackage{multirow}

\usetikzlibrary{quotes,arrows.meta}

\newcommand{\hm}{Hyundai Motor Company\xspace}
\newcommand{\hmr}{R\&D Division at Hyundai Motor Company\xspace}

\newcommand{\train}{$T_r$\xspace}
\newcommand{\test}{$T_t$\xspace}

\newcommand{\tradd}{$T'_r$\xspace}
\newcommand{\teadd}{$T'_t$\xspace}

\def\genbox#1#2#3#4#5#6{% #1=0/1, #2=color, #3=shape, #4=raise, #5=width, #6=width/2
    \leavevmode\raise#4bp\hbox to#5bp{\vrule height#5bp depth0bp width0bp
    \pdfliteral{q .5 w \csname #2COLOR\endcsname\space RG
                       \csname #3PDF\endcsname{#5}{#6} S Q
             \ifx1#1 q \csname #2COLOR\endcsname\space rg
                       \csname #3PDF\endcsname{#5}{#6} f Q\fi}\hss}}

\AtBeginDocument{%
  \providecommand\BibTeX{{%
    \normalfont B\kern-0.5em{\scshape i\kern-0.25em b}\kern-0.8em\TeX}}}

\setcopyright{acmcopyright}
\copyrightyear{2020}
\acmYear{2020}
\begin{document}

\title[Reducing DNN Labelling Cost using Surprise Adequacy: An Industrial Case Study for Autonomous Driving]{Reducing DNN Labelling Cost using Surprise Adequacy:\\An Industrial Case Study for Autonomous Driving}
%\title{Generative Model Inference for Search Based Test Data Generation}

\author{Jinhan Kim}
\affiliation{%
 \institution{School of Computing, KAIST}
 \city{Daejon}
 \country{Republic of Korea}
}
\email{jinhankim@kaist.ac.kr}

\author{Jeongil Ju}
\affiliation{%
 \institution{Hyundai Motor Group}
 \city{Seoul}
 \country{Republic of Korea}
}
\email{littlespice@hyundai.com}

\author{Robert Feldt}
\affiliation{%
  \institution{Chalmers University}
  \city{Gothenburg}
  \country{Sweden}
}
\email{robert.feldt@chalmers.se}

\author{Shin Yoo}
\affiliation{%
 \institution{School of Computing, KAIST}
 \city{Daejon}
 \country{Republic of Korea}
}
\email{shin.yoo@kaist.ac.kr}


\makeatletter
\def\@copyrightspace{\relax}
\makeatother

\begin{abstract}
Deep Neural Networks (DNNs) are rapidly being adopted by the
automotive industry, due to their impressive performance in tasks that are
essential for autonomous driving. Object segmentation is one such task: its aim
is to precisely locate boundaries of objects and classify the identified
objects, helping autonomous cars to recognise the road environment and
the traffic situation. Not only is this task safety critical, but
developing a DNN based object segmentation module presents a set of challenges
that are significantly different from traditional development of safety
critical software. The development process in use consists of multiple iterations of
data collection, labelling, training, and evaluation. Among these stages,
training and evaluation are computation intensive while data collection and
labelling are manual labour intensive. This paper shows how
development of DNN based object segmentation can be improved by exploiting the
correlation between Surprise Adequacy (SA) and model performance. The
correlation allows us to predict model performance for inputs without
manually labelling them. This, in turn, enables understanding of model
performance, more guided data collection, and informed decisions about further
training. %, all without costly manual labelling.
%We demonstrate how SA
%can help improving the development process of DNN based object segmentation.
In our industrial case study the technique allows cost savings
of up to 50\% with negligible evaluation inaccuracy. Furthermore, engineers can
trade off cost savings versus the tolerable level of inaccuracy depending
on different development phases and scenarios.

%\sy{I want to include a specific number here, like
%X\% labelling cost saving at Y\% hit at evaluation accuracy, but there is
%no specific trade-off point that really stands out... shall we say ``up to 50\%
%with negligible evaluation inaccuracy'', or is that too much? ``depending on evaluation metric''?}
\end{abstract}


\maketitle

\section{Introduction}
\label{sec:intro}

Machine Learning technologies such as Deep Neural Networks (DNNs) are
increasingly used as components in complex software and industrial systems
deployed to customers. While much research has focused on how to improve and
then test the performance and robustness of these components their increased
use poses a number of additional challenges to software engineers and
managers. For example, while training and retraining after refinements (of
Manuscript data or model setup) of the DNNs are not primarily labour but compute intensive
it can take considerable time and thus delay the development process. There
are also considerable costs involved in collecting, ensuring the quality of,
and then labelling the data to enable supervised training. A fundamental
challenge is also to judge how much additional training should be done and on
which data.

Individual solutions to several of these problems have been proposed. For example, so-called
active learning has been proposed to reduce labelling costs, but is often relatively
complex involving ensembles of networks~\cite{Beluch2018pt} or limits
the type of deep networks that can be used~\cite{Gal2017rq}.
Here, we investigate if a simple metric for quantifying how surprising
an input is to a DNN, can be used to support several of these real-world
engineering challenges in an industrial setting. In particular, we focus on a key step in the
processing pipeline of an autonomous, so-called self-driving, car: the
segmentation of images into separate objects for later processing of dangers
as well as planning of steering and actions. This is a complex, embedded, and
a real-time system encompassing both hardware and software and needing to use
state of the art DNN technologies. While much DNN research has focused on
image recognition and classification and later on object detection, semantic
segmentation of images is harder still. To ensure that a DNN-based component of
the car can perform this task robustly is a major challenge in the automotive industry
and, as such, can serve as a testbed for the supporting technology we propose. Not
only is the task technically very challenging, it also has one of the highest
labelling costs.

The approach we have evaluated with our industrial partner exploits our
previously proposed Surprise Adequacy (SA) metric for estimating how surprising a
new input is to a DNN~\cite{Kim2019aa}. This metric was introduced primarily as a
test adequacy criterion, i.e. a way to select test cases and evaluate if a DNN
is sufficiently robust and of high enough quality. However, the actual measure
quantifies how different a new input is to the ones the network has already
seen. Since those are the ones the network has been trained on, and thus where
it should perform well, we can further exploit the metric to guide labelling,
training and retraining scenarios. Here we evaluate this potential in a real,
industrial setting and based on the key challenges identified by the
practitioners in the company.

The rest of the paper is organised as follows. \Cref{sec:objseg} introduces the
DNN based semantic segmentation and further details its challenges. \Cref{sec:sa}
briefly describes the Surprise Adequacy (SA) from our previous work, and explains
how it is applied to the semantic segmentation DNN models; it also introduces
a new type of Surprise Adequacy metric called Mahalanobis Distance based SA.
\Cref{sec:experimentaldesign} describes the datasets we use, and presents
our research questions. \Cref{sec:evaluation} presents and discusses the
evaluation results of our proposed technique. \Cref{sec:threats} lays out
threats to validity, and \Cref{sec:relatedwork} discusses the related work.
Finally, \Cref{sec:conclusion} concludes with a reference to future work.

\begin{figure}[h]
\begin{subfigure}[b]{0.45\linewidth}
\includegraphics[width=35mm]{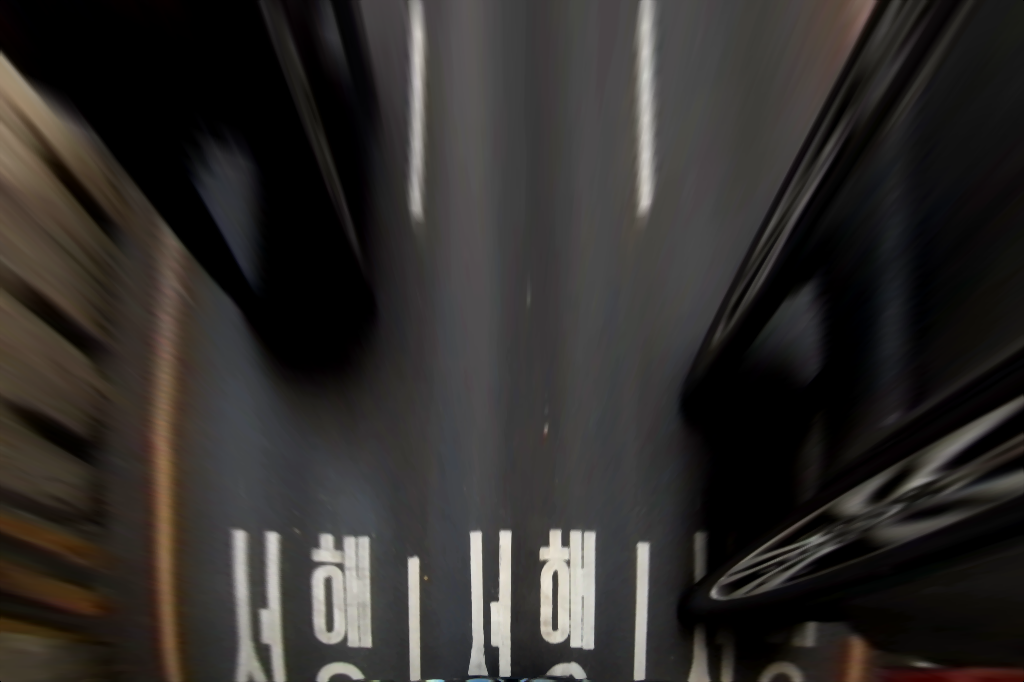}
\caption{Input Image\label{fig:ex_input}}
\end{subfigure}
\begin{subfigure}[b]{0.45\linewidth}
\includegraphics[width=35mm]{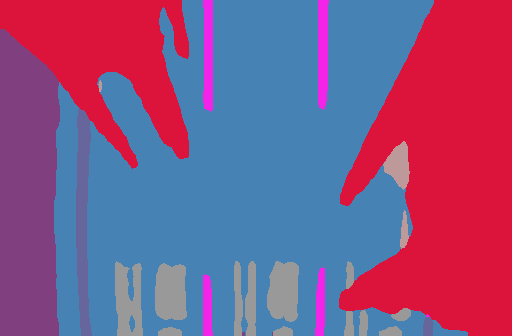}
\caption{Output Segmentation\label{fig:ex_output}}
\end{subfigure}
\Description{Example input and output of semantic segmentation}
\caption{An input and output of semantic segmentation
\label{fig:example_segmentation}}
\end{figure}

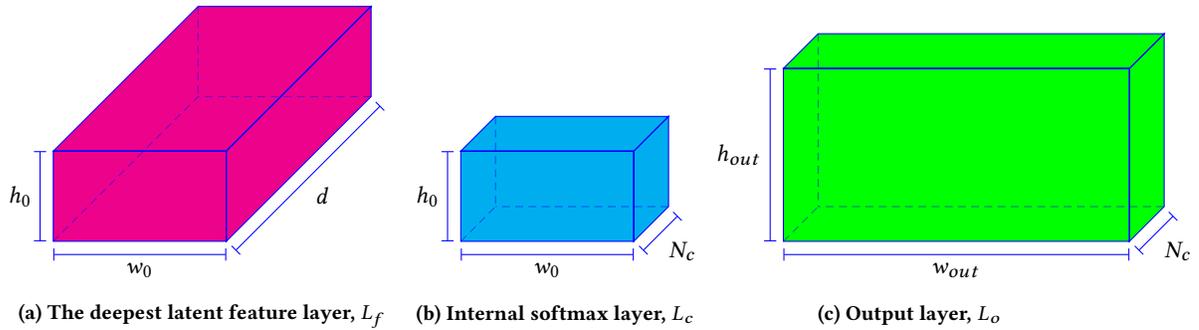
\begin{figure*}[!ht]
\begin{subfigure}[b]{0.3\textwidth}
\begin{tikzpicture}[every edge quotes/.append style={auto, text=black}]
  \pgfmathsetmacro{\cubex}{2.3}%
  \pgfmathsetmacro{\cubey}{1.2}%
  \pgfmathsetmacro{\cubez}{5}%
  \draw [draw=blue, every edge/.append style={draw=blue, densely dashed, opacity=.5}, fill=magenta]
    (0,0,0) coordinate (o) -- ++(-\cubex,0,0) coordinate (a) -- ++(0,-\cubey,0) coordinate (b) edge coordinate [pos=1] (g) ++(0,0,-\cubez)  -- ++(\cubex,0,0) coordinate (c) -- cycle
    (o) -- ++(0,0,-\cubez) coordinate (d) -- ++(0,-\cubey,0) coordinate (e) edge (g) -- (c) -- cycle
    (o) -- (a) -- ++(0,0,-\cubez) coordinate (f) edge (g) -- (d) -- cycle;
  \path [every edge/.append style={draw=blue, |-|}]
    (b) +(0,-5pt) coordinate (b1) edge ["$w_{0}$"'] (b1 -| c)
    (b) +(-5pt,0) coordinate (b2) edge ["$h_{0}$"] (b2 |- a)
    (c) +(3.5pt,-3.5pt) coordinate (c2) edge ["$d$"'] ([xshift=3.5pt,yshift=-3.5pt]e)
    ;
\end{tikzpicture}
\caption{The deepest latent feature layer, $L_f$\label{fig:layer_a}}
\end{subfigure}
\begin{subfigure}[b]{0.22\textwidth}
\begin{tikzpicture}[every edge quotes/.append style={auto, text=black}]
  \pgfmathsetmacro{\cubex}{2.3}%
  \pgfmathsetmacro{\cubey}{1.2}%
  \pgfmathsetmacro{\cubez}{1.2}%
  \draw [draw=blue, every edge/.append style={draw=blue, densely dashed, opacity=.5}, fill=cyan]
    (0,0,0) coordinate (o) -- ++(-\cubex,0,0) coordinate (a) -- ++(0,-\cubey,0) coordinate (b) edge coordinate [pos=1] (g) ++(0,0,-\cubez)  -- ++(\cubex,0,0) coordinate (c) -- cycle
    (o) -- ++(0,0,-\cubez) coordinate (d) -- ++(0,-\cubey,0) coordinate (e) edge (g) -- (c) -- cycle
    (o) -- (a) -- ++(0,0,-\cubez) coordinate (f) edge (g) -- (d) -- cycle;
  \path [every edge/.append style={draw=blue, |-|}]
    (b) +(0,-5pt) coordinate (b1) edge ["$w_{0}$"'] (b1 -| c)
    (b) +(-5pt,0) coordinate (b2) edge ["$h_{0}$"] (b2 |- a)
    (c) +(3.5pt,-3.5pt) coordinate (c2) edge ["$N_c$"'] ([xshift=3.5pt,yshift=-3.5pt]e)
    ;
\end{tikzpicture}
\caption{Internal softmax layer, $L_c$\label{fig:layer_b}}
\end{subfigure}
\begin{subfigure}[b]{0.3\textwidth}
\begin{tikzpicture}[every edge quotes/.append style={auto, text=black}]
  \pgfmathsetmacro{\cubex}{4.6}%
  \pgfmathsetmacro{\cubey}{2.3}%
  \pgfmathsetmacro{\cubez}{1.2}%
  \draw [draw=blue, every edge/.append style={draw=blue, densely dashed, opacity=.5}, fill=green]
    (0,0,0) coordinate (o) -- ++(-\cubex,0,0) coordinate (a) -- ++(0,-\cubey,0) coordinate (b) edge coordinate [pos=1] (g) ++(0,0,-\cubez)  -- ++(\cubex,0,0) coordinate (c) -- cycle
    (o) -- ++(0,0,-\cubez) coordinate (d) -- ++(0,-\cubey,0) coordinate (e) edge (g) -- (c) -- cycle
    (o) -- (a) -- ++(0,0,-\cubez) coordinate (f) edge (g) -- (d) -- cycle;
  \path [every edge/.append style={draw=blue, |-|}]
    (b) +(0,-5pt) coordinate (b1) edge ["$w_{out}$"'] (b1 -| c)
    (b) +(-5pt,0) coordinate (b2) edge ["$h_{out}$"] (b2 |- a)
    (c) +(3.5pt,-3.5pt) coordinate (c2) edge ["$N_c$"'] ([xshift=3.5pt,yshift=-3.5pt]e)
    ;
\end{tikzpicture}
\caption{Output layer, $L_o$\label{fig:layer_c}}
\end{subfigure}
\Description{}
\caption{Dimensions of the object segmentation output layer. Each point in the
$w_0$ by $h_0$ plane in \cref{fig:layer_a} is associated with $d$ latent
features. These are aggregated into $N_c$ softmax scores in the softmax layer
shown in \cref{fig:layer_b}. Finally, these scores are upsampled to the output
dimension, $w_{out}$ by $h_{out}$ in the output layer shown in
\cref{fig:layer_c}.\label{fig:typical_output_layer}}
\end{figure*}

\section{Semantic Segmentation for Autonomous Driving}
\label{sec:objseg}

Object recognition techniques for images can be categorised based on the number of
objects being recognised, as well as the precision of the location information
being extracted. Image classification, which is the subject of many existing
work on DNN development and testing~\cite{Pei2017qy,Tian2018aa,Kim2019aa}, simply aims to put
a single label on the entire input image, which often contains a single or main
object. Object detection, on the other hand, accepts images that contain
multiple objects, and aims to put not only labels but also bounding boxes to
each object in the image~\cite{Feng2020mz}. Finally, semantic segmentation
aims to partition the input image into meaningful parts by labelling each
\emph{pixel} in the image. \Cref{fig:example_segmentation} shows a semantic
segmentation example with an input on the left and its output on the right.

Semantic segmentation is a critical task for autonomous driving, as it allows
the vehicle to correctly recognise the traffic scene around itself thus enabling
analysis, detection of dangers, as well as being the basis for planning and action~\cite{
Siam2018dz}. This section briefly describes the basic principles of semantic
segmentation, and the challenges of developing a DNN based semantic
segmentation module for autonomous driving in an industrial setting.

\subsection{DNN Based Object Segmentation}
\label{sec:dnnobjseg}

A fundamental principle shared by all existing approaches is that semantic
segmentation of an image consists of many instances of the pixel
classification problem. In image classification, we typically learn low level
features placed across the entire image using convolution, and subsequently
combine these to reach a classification result~\cite{Simonyan2014aa,
Szegedy2015pt}. In semantic segmentation, we need to learn and represent multiple low level features
for each pixel, resulting in more complicated architectures.

\Cref{fig:typical_output_layer} shows the three deepest (sets of) layers of the semantic
segmentation DNN model trained by the \hmr (details irrelevant to our
discussion
are left out). The model can segment objects in $N_c$ different classes, takes
an input image of $w_{in}$ by $h_{in}$, and
returns the segmentation results in output images of size $w_{out}$ by
$h_{out}$ (both $w_{in}$ and $w_{out}$ are set to 512, and $h_{in}$ and
$h_{out}$ are set to 336). The DNN model learns the latent features used
for segmentation via the use of various convolution
layers~\cite{Szegedy2015pt}. With the use of convolution layers, the internal
representation of the input image is significantly smaller than the input and
output image: let us denote the size of the internal representation by $w_0$
by $h_0$ (which is 64 by 42 in the case of the model we study). Note that the use of the smaller
internal representation is not specific to the model studied by us, and used
by other existing semantic segmentation techniques~\cite{Wu2018oz,Mehta2018pi}.

% \rf{See if you can clarify the previous sentence, there are quite a few things going on here. Maybe clarify that $w_0$ and $h_0$ are smaller than the input image by giving the dimensions also of the latter?}
% \sy{I tried to make it more detailed - it is a but "convoluted" (pun intended) because I did not want to reveal the specific dimension, as it may be connected to existing models, but I guess that is unavoidable...}

\Cref{fig:layer_a} shows the
latent feature layer, $L_f$, each \emph{pixel} of which is associated with $d$
latent features. The next layer, $L_c$, shown in \cref{fig:layer_b}, performs
the pixel-level classification by computing the softmax scores for $N_c$ class
labels. Finally, the output layer, $L_o$, upsamples $L_c$ to the output image
size, $w_{out}$ by $h_{out}$. Let $L_o(x, y)$ be the softmax score vector,
$z$, of length $N_c$: the class label of the pixel is $argmax{(z)}$. The
result of the segmentation can be represented as an image of size $w_{out}$ by
$h_{out}$, in which each pixel has the specified colour of its class label.
Note that, while the layers shown in \Cref{fig:typical_output_layer} are
specific to the industrial model we study here, a similar structure can be found in other
semantic segmentation models.

\subsection{Labelling Cost for Semantic Segmentation}
\label{sec:cost}

\begin{figure}[ht]
\includegraphics[width=60mm]{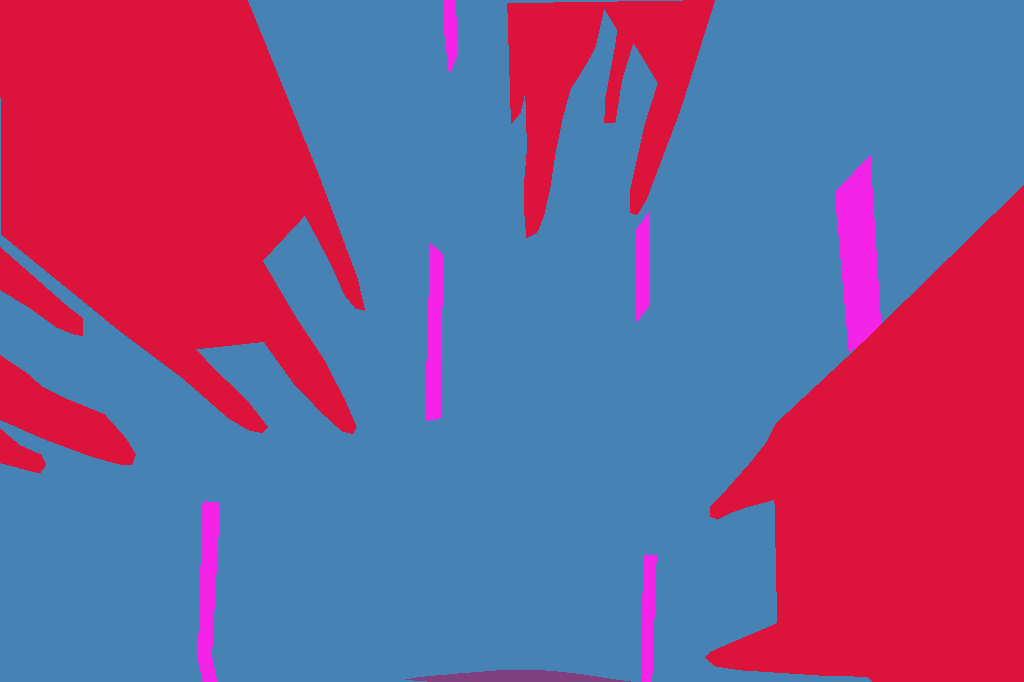}
\Description{Example manual labelling}
\caption{Example manual labelling\label{fig:example_labelling}}
\end{figure}

All supervised learning requires manually labelled datasets, which is costly
to build. Semantic segmentation requires particularly expensive labelling:
unlike image classification, for which the act of labelling is entering the
name of the object in the given image, semantic segmentation requires pixel
classification via the act of colouring different areas occupied by different
types of objects. \Cref{fig:example_labelling} shows an example manual labelling. The task is both laborious and time consuming.

While the exact internal cost of labelling at \hm will remain confidential, the
scale
of the problem can be conveyed by looking at the publicly available data
labelling services. Semantic segmentation is the most expensive image-based
labelling task provided by Google Cloud.\footnote{Refer to \url{
https://cloud.google.com/ai-platform/data-labeling/pricing\#labeling_costs}}
Each segment labelled by an individual worker becomes a \emph{unit}: each
month, the first 50,000 units are available at 0.87 USD per unit, and the
following 950,000 units at 0.85 USD per unit.

In the dataset we study, each image typically contains five to ten
segments; if we require at least three individual workers per image for
robustness, each image will consume 15 to 30 units. However, note that our
``images'' are actually frames from video that is captured during driving.
Assuming the standard 24fps (frames per second), one second of video results
in 360 to 720 units, costing anywhere roughly between 300 and 600 USD. Even if
we consider more attractive bulk pricing, the cost is clearly non-trivial; any
savings that can be made without detrimental effects on the accuracy of the training
process will be important. The primary goal of this study is to see whether
SA can help to reduce this cost by acting as a surrogate measure for model
performance. If there is a strong correlation between SA and the model
performance, we can use SA to prioritise inputs that must be manually
labelled: inputs with lower SA can either be skipped, or be given lower
priority in labelling urgency, as the model is more likely to handle them
correctly.

\subsection{Guiding Iterative Retraining}
\label{sec:retraining}

Unlike traditional software system whose code is written by human developers,
DNNs are trained from training data~\cite{Yoo2019aa}. Consequently, if the
performance of a trained DNN model is not satisfactory, it has to go through a
series of retraining instead of patching. To improve the model performance,
it is widely accepted that the training data for retraining should be curated
carefully~\cite{Ma2018gf,Kim2019aa}. We posit that SA can improve the
effectiveness of iterative retrainings, by enabling the engineers to choose
inputs that are sufficiently diverse from the existing training data.

% First, if our correlation-based hypothesis holdes
% and can reduce labelling cost, it also means that we can choose inputs for
% retraining without necessarily labelling them manually: this will improve the
% efficiency of the
% retraining process, both time- and cost-wise. Second, SA may be able to
% provide a finer control over the composition of retraining data. Without the
% guidance from SA, one can only add more inputs from the type of which the
% model does not perform well; with the guidance of SA, it is possible to gain
% more insights about input diversity and make informed selection.

% \rf{Not clear to me what we are saying above, needs to be clarified. I suggest we make a clear difference between saving by not having to label, or savings since we can select the inputs that would improve quality the most. The latter might still require labelling though. I didn't rewrite here since not clear you mean the same things as I...}
% \sy{you're right, my message was meddled, I simplified the whole thing}

\section{Surprise Adequacy}
\label{sec:sa}

This section describes how Surprise Adequacy (SA)~\cite{Kim2019aa} has been
applied to our semantic segmentation DNN. The basic idea of SA is that
distances between the internal values of a neural network, i.e. the values
calculated in each node of the network when activated by an input, carry
information about how similar these inputs are.

Thus, if we simply `run' the network on an input and save all the node
activation values in a (long) vector, we can compare such vectors and thus
quantify how distant inputs or groups of inputs are from each other. If the
vector has a large distance to the vectors seen by the network during
training, the internal computations of the network are different, and we
consider the input causing the vector to be surprising. From a software
engineering perspective this is natural, since the node activation values
correspond to state values during the execution of a software function. Since
testing on non-surprising inputs is not likely to uncover incorrect DNN
behaviour, a goal of DNN testing becomes to find inputs that are adequately
surprising. Hence, the name.

In SA terminology, these vectors of node activation values are called
Activation Traces (ATs) and different distances between them can be used for
different purposes in DNN training, testing, and engineering~\cite{Kim2019aa}.
In general terms, we can do one-to-one SA, by comparing individual ATs to each
other, one-to-many SA to quantify how distant one AT is to a set of others, or
we can do many-to-many SA by comparing two groups of ATs to each other. The
original SA paper~\cite{Kim2019aa}, proposed the one-to-one distance-based SA,
DSA, to explore class boundaries for DNN classification tasks and the
one-to-many likelihood-based SA (LSA) as a general SA metric for both
regression and classification. In the industrial setting explored here we
primarily need a one-to-many distance metric and thus focus on LSA. However,
LSA is computationally costly so for industrial applicability we searched for
ways of making it more efficient to compute.

Below, we first describe how we extract Activation Traces (ATs) from the DNNs
used for semantic segmentation in autonomous cars, and then present the actual
computation of SA values we use in this study. In particular, we describe a
new type of SA, called Mahalanobis Distance Based SA (MDSA) with preferable
scaling properties.

\subsection{Activation Traces for Object Segmentation}
\label{sec:ats}

Since semantic segmentation is essentially classification of each pixel in the
input image, the output is also an image with the same dimension as the input.
However, as can be seen in \Cref{fig:typical_output_layer}, the semantic
segmentation DNN we studied performs classification first (when going from
$L_f$ to $L_c$) and upsamples the \emph{results} of the classification to the
output dimension (when going from $L_c$ to $L_o$). This poses a problem for
the SA analysis due to its nature. The surprise we want to measure essentially
captures how similar the features of an unseen pixel are to the features of
pixels in the training data. If we do not consider pixels and the
corresponding features separately based on their real class labels, the
resulting surprise will end up capturing how similar an unseen pixel is to all
the pixels in the training data: this is likely to be extremely noisy.
Consequently, to compute SA per class label, we need to map the classification
features (a vector of length $d$) to a label (the value of a pixel in the provided label image).
However, the feature vector of length $d$ only exists for pixels in $w_0$ by
$h_0$ plane, with no one to one mapping from feature vectors to pixels in the
output: we have $w_{out} \times h_{out}$ labels, but only $w_{0} \times h_{0}$
feature vectors (with $w_{0} < w_{out}$, etc).
% \rf{Not clear above and I'm not sure what the argument is here. Is the problem that even though a particular value in a SA AT always represent the same node in a network this node might be used to represent different "features" on differen invocations/inputs? This is not clear from the text IMHO.}
% \sy{the problem is that we have fewer feature vectors than actual labels, since the classification happens in an "image" that is smaller than the output - then the classification result is upsampled to match the output dimension}
% \rf{Ok, this is more clear now.}

\begin{figure}[ht]
\begin{subfigure}[b]{0.4\linewidth}
\begin{tikzpicture}[every edge quotes/.append style={auto, text=black}]
  \pgfmathsetmacro{\cubex}{1.65}%
  \pgfmathsetmacro{\cubey}{0.6}%
  \pgfmathsetmacro{\cubez}{2.5}%
  \draw [draw=blue, every edge/.append style={draw=blue, densely dashed, opacity=.5}, fill=magenta]
    (0,0,0) coordinate (o) -- ++(-\cubex,0,0) coordinate (a) -- ++(0,-\cubey,0) coordinate (b) edge coordinate [pos=1] (g) ++(0,0,-\cubez)  -- ++(\cubex,0,0) coordinate (c) -- cycle
    (o) -- ++(0,0,-\cubez) coordinate (d) -- ++(0,-\cubey,0) coordinate (e) edge (g) -- (c) -- cycle
    (o) -- (a) -- ++(0,0,-\cubez) coordinate (f) edge (g) -- (d) -- cycle;
  \path [every edge/.append style={draw=blue, |-|}]
    (b) +(0,-5pt) coordinate (b1) edge ["$w_{0}$"'] (b1 -| c)
    (b) +(-5pt,0) coordinate (b2) edge ["$h_{0}$"] (b2 |- a)
    (c) +(3.5pt,-3.5pt) coordinate (c2) edge ["$d$"'] ([xshift=3.5pt,yshift=-3.5pt]e)
    ;
\end{tikzpicture}
\caption{The deepest latent feature layer, $L_f$\label{fig:layer_feature}}
\end{subfigure}
\begin{subfigure}[b]{0.45\linewidth}
\begin{tikzpicture}[every edge quotes/.append style={auto, text=black}]
  \pgfmathsetmacro{\cubex}{2.3}%
  \pgfmathsetmacro{\cubey}{1.15}%
  \pgfmathsetmacro{\cubez}{2.5}%
  \draw [draw=blue, every edge/.append style={draw=blue, densely dashed, opacity=.5}, fill=pink]
    (0,0,0) coordinate (o) -- ++(-\cubex,0,0) coordinate (a) -- ++(0,-\cubey,0) coordinate (b) edge coordinate [pos=1] (g) ++(0,0,-\cubez)  -- ++(\cubex,0,0) coordinate (c) -- cycle
    (o) -- ++(0,0,-\cubez) coordinate (d) -- ++(0,-\cubey,0) coordinate (e) edge (g) -- (c) -- cycle
    (o) -- (a) -- ++(0,0,-\cubez) coordinate (f) edge (g) -- (d) -- cycle;
  \path [every edge/.append style={draw=blue, |-|}]
    (b) +(0,-5pt) coordinate (b1) edge ["$w_{out}$"'] (b1 -| c)
    (b) +(-5pt,0) coordinate (b2) edge ["$h_{out}$"] (b2 |- a)
    (c) +(3.5pt,-3.5pt) coordinate (c2) edge ["$d$"'] ([xshift=3.5pt,yshift=-3.5pt]e)
    ;
\end{tikzpicture}
\caption{Activation Trace layer, $L_{AT}$\label{fig:layer_extraction}}
\end{subfigure}
\Description{}
\caption{Activation Trace (AT) extraction for semantic segmentation\label{fig:at_layer}}
\end{figure}
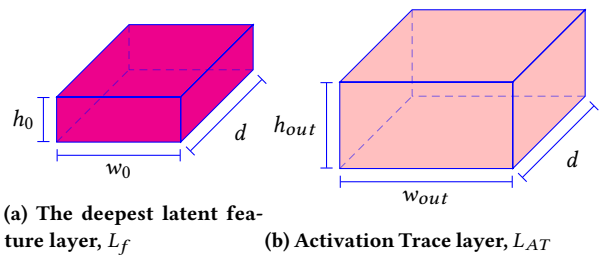

\Cref{fig:at_layer} shows how we circumvent this problem by modifying the
model architecture specifically for AT extraction. The semantic segmentation
model performs classification in a smaller dimension ($w_{0}$ by $h_{0}$) and
upsamples the resulting softmax scores; we, instead, upsample the features
directly so that we can have one to one mapping between features and pixel
labels. The same upsampling algorithm that is used when going from $L_c$ to
$L_o$ has been applied to $L_f$ to obtain the \emph{instrumentation} layer, $L_{AT}$. To extract ATs, we simply store $L_{AT}$ during execution. The AT vector
for a pixel at $(w, h)$ in the model output of size $w_{out} \times h_{out}$
is the feature vector of length $d$ at $(w, h)$ in $L_{AT}$.
%\rf{Yes, this works better now}.

Note that, for each pixel, the feature (AT) vector of length $d$ that we extract
can be thought of as the activation values of the deepest, final fully
connected layer before the softmax layer of the single pixel classifier DNN.
Also, given that the classification takes place at the pixel level, we are no
longer bound by \emph{images} as the unit of analysis. Instead, each
individual pixel from a specific class label will be the unit of analysis.
% \rf{Please clarify above what the actual AT vector is finally made up of. Just stacking the vectors for each class after each other? $L_AT$ not seen in the SA formulas below so quite confusing.}

\subsection{Likelihood Based SA}
\label{sec:lsa}

Likelihood Based SA (LSA) uses Kernel Density Estimation (KDE) to summarise
the ATs. Let $P_c$ be the set of pixels that belong to class $c$. Using the
notation from our previous work~\cite{Kim2019aa}, let $A_{L_f}(P_c)$ be the
set of individual ATs, $\alpha_{L_f}(p)$, for pixel $p$ in $P_c$. Given a new
input pixel, $x$, we first perform the KDE:

\begin{equation}
\label{eq:lsa_kde}
\hat{f}(x)=\frac{1}{|A_{L_{f}}(P_c)|} \sum_{x_{i} \in P_c} K_{H}\left(\alpha_{L_f}(x)-\alpha_{L_f}\left(x_{i}\right)\right)
\end{equation}

Here, $H$ denotes the bandwidth matrix, and $K$ is a Gaussian kernel function.
Based on KDE, we compute the Likelihood based Surprise Adequacy of the new
input $x$ as:

\begin{equation}
LSA(x) = -\log(\hat{f}(x))
\end{equation}

\subsection{Mahalanobis Distance Based SA}

% \sy{Robert: please refer to the TOSEM extension}
% \rf{Please check if the description below matches your use of MD. Can we refer to the TOSEM extension already here in submission to FSE? Overlapping submission processes, I mean...}
%  \sy{I simply meant for you to check what you already wrote for TOSEM :)}
While LSA was shown~\cite{Kim2019aa} to be an effective way to quantify how far
an activation trace is from a set of other traces, its performance suffers as
this set becomes larger. The problem is that the KDE requires the new
activation trace to be compared to all the existing ones in the set, e.g. the
summation in Equation~\ref{eq:lsa_kde} above is over ATs for all inputs in set
$P_c$. This is unsuitable for typical, industrial use cases where the training
data may contain a very large number of inputs, leading to significantly many
activation traces of non-trivial lengths. In summary, $LSA$, despite being
effective, can have unfavourable scaling properties for large-scale,
industrial application.
%\sy{I think we address the "long length" of ATs by being selective with layers. Here the problem is that we have too many. I had a go at rewriting it above.}

For the industrial setting of this paper, we addressed this shortcoming by
seeking a summary of the activation traces seen so far. A natural choice is to
use the Mahalanobis distance~\cite{De-Maesschalck2000cj}. This is a
generalisation of the Euclidean distance that takes correlation in a dataset
into account and can measure the distance between a point (here: an activation
trace as a vector) and a distribution (here: the distribution of previously
seen traces we want to compare the new point to). Its use in software
engineering is rare but it was part of a method for software defect prediction~
\cite{Liparas2012ys}; in the context of DNN models, it has been
successfully applied to detect out-of-distribution inputs~\cite{Lee2018hf}.
Here, instead of having to loop over all the
activation traces for points in $P_c$ we can precalculate the mean, $\mu_c$
and covariance matrix, $S_c$ and can then calculate the Mahalanobis Distance
based Surprise Adequacy (MDSA) for an input $x$:

\begin{equation}
\label{eq:mdsa}
MDSA(x) = \sqrt{(\alpha_{L_f}(x) - \mu_c)^{T} S_{c}^{-1} (\alpha_{L_f}(x) - \mu_c)}
\end{equation}

This measures the distance from an AT to the centre of the
previous ATs while taking the amount of variation among the
latter, in the direction to the new point, into account. Since the inversion
of the covariance matrix can be cached for the set $P_c$, calculating the MDSA
involves only five elementary, linear algebra operations. This leads to
considerable speedups in practice.

\section{Experimental Design}
\label{sec:experimentaldesign}

This section describes our experimental design via the datasets we use as well
as the research questions we study.

\subsection{Datasets}
\label{sec:datasets}

\Cref{tab:datasets} shows the four different datasets used in this paper.
Initially, training and test datasets of small scale, \train and \test, are
used to study the feasibility of SA analysis (RQ1) and to conduct the
labelling cost study (RQ2). Subsequently, an additional and larger dataset
from a subsequent data collection campaign has been made available. We
conduct the retraining guidance study (RQ3) using the training (\tradd) and
test (\teadd) datasets from this second batch.

\begin{table}[ht]
\caption{Four different datasets used in this paper\label{tab:datasets}}
\scalebox{0.80}{
\begin{tabular}{lrrll}
\toprule
Name & Size & Classes & Description & RQs \\ \midrule
\train & 16,549 & 12 & The 1st training data  & \multirow{2}{*}{RQ1, 2}\\
\test  &  2,186 & 12 & The 1st test data \\ \midrule
\tradd & 60,532 & 14 & The 2nd training data & \multirow{2}{*}{RQ3}\\
\teadd & 10,000 & 14 & The 2nd test data \\
% \teadd & \fixme{?} & The additional test data pool \\
\bottomrule
\end{tabular}}
\end{table}

Manually generated segmentation labels have also been provided for all four
datasets. There are 14 segmentation classes, which are shown in
\Cref{tab:classes}. Both \train and \test are initially labelled with 12 classes,
while \tradd and \teadd use two additional, more fine grained class labels
(vehicles are further segmented into bodies and wheels, and external road
structures are segmented from void regions).
While lanes and road markers are divided into separate
classes for easier segmentation based on features such as different colours
and shapes, we will treat them as only two semantic class groups: class group
123 (lanes) and class group 45678 (road markers, i.e., drawings and writings
on the road). The void class represents the dark areas outside the road. In
\train and \test, all areas outside the road have been labelled as void.
Consequently, the areas labelled as void lack consistent features and we
expect noisy results (as no meaningful segmentation is possible based on their
own features). In \tradd and \teadd, if there are any visible structures
outside, they have been labelled separately as general structures. Finally,
class 9 has been excluded from our analysis, as \test contained only 32 images
with toll gate markers.

\begin{table}[ht]
\caption{Segmentation Classes\label{tab:classes}}
\scalebox{0.75}{
\begin{tabular}{cl|cl}
\toprule
Class & Description & Class & Description \\ \midrule
0 & Void                     & 7  & Road Marker (Numbers)   \\
1 & Lanes (White)            & 8  &  Road Marker (Crosses)  \\
2 & Lanes (Blue)             & 9  &  Toll Gate Marker       \\
3 & Lanes (Yellow)           & 10 &  Vehecles               \\
4 & Road Marker (Arrows)     & 11 &  Road Area              \\
5 & Road Marker (Shapes)     & 12 &  Vehicle Wheels (\tradd, \teadd only) \\
6 & Road Marker (Characters) & 13 &  General Structures (\tradd, \teadd only)\\
\bottomrule
\end{tabular}}
\end{table}

The images in all of these datasets are frames of various segments of video,
which has been recorded using fisheye view cameras mounted on the data
collection vehicle~\cite{Yurtsever2019tx}. We use \texttt{ffmpeg}\footnote{\url
{https://ffmpeg.org}} library to extract video frames into bitmap images.
Video segments are not necessarily consecutive and include various driving
conditions, such as different road conditions and time of day, etc.

We train state-of-the-art semantic segmentation models using datasets
described above. All models are implemented using Python and
PyTorch~\cite{Paszke2019la}. The maximum epoch is set to 300, batch size to
128, learning rate to $10^{-5}$, weight decay to $0.0005$: all hyperparameters
have been empirically tuned by the \hmr.

% \fixme{we do not say anything about the model - should we list the
% hyperparameters?}

% \rf{Maybe just state the main ones? We don't have a lot of space but it will be requested by reviewers I think...}

\subsection{Research Questions}
\label{sec:rqs}

We ask the following three research questions:\\

\noindent\textbf{RQ1. Feasibility:} the first research question is a sanity
check that the AT extraction described in \Cref{sec:ats}, and
the SA analysis that follows, work as expected. Does the model perform worse with more surprising inputs?
We answer RQ1 by plotting, and by reporting Spearman correlation between,
the SA values and the model performance metrics. The Spearman correlation is a
natural choice, in this case, since it does not make assumptions about the
underlying distribution of data and can better handle non-linearities that are
to be expected in DNNs.

To capture the model performance, we use Intersection over Union (IoU);
the standard evaluation metric for semantic segmentation. IoU is the ratio
between the intersection of the predicted segment (i.e., region) and the label
segment and the union of the two segments. When the predicted segment is
exactly the same shape as the labelled segment, IoU becomes 1.0. The less they overlap
the closer to 0.0 the IoU becomes. In practice, we
compute the IoU for a specific segment class by considering all pixels that
belong to that class, instead of computing IoU for each independent segment.
If objects from $n$ classes are present in a single image,
we get $n$ different IoU values.\\
% \rf{Clarify if we thus get multiple IoU values per image (since there are multiple objects and classes).}
% \sy{done!}

\noindent\textbf{RQ2. Cost Efficiency:} the second research question concerns
how much labelling cost can be saved by using SA values as a guide. To study the trade off
between labelling cost saving and the resulting inaccuracy in model
evaluation, we simulate a scenario in which we do not label the $x$\% of
images with the lowest SA. If SA correlates well with model performance the images with the
lowest SA values can be expected to add very little performance even if they were labelled.
RQ2 is answered by plotting the inaccuracy from not
labelling those images against $x$.

By not labelling images with low SA values, we explicitly accept some
inaccuracy in model evaluation. We measure the model \emph{inaccuracy} for the
skipped images by reporting the complement of IoU, i.e., 1 - IoU.
Additionally, we also adopt the standard accuracy metric: for this, we
consider the segmentation for a specific class in an entire image to be \emph{
problematic} only if the IoU for the given class is below a pre-determined
threshold. This choice was made based on the experience from the engineers at the company.
Suppose we set the threshold to 0.5 and consider vehicles in the
given image: we will consider a segmentation to be problematic only when the
predicted vehicle segments collectively cover less than 50\% of the label
vehicle segments. By not labelling some images, we are effectively accepting
the predicted segmentation as correct. Consequently, the inaccuracy simply
becomes the ratio of unlabelled images whose IoU is below the given threshold.
We use IoU threshold values from 0.5 to 0.9 with the interval of 0.1.\\

\noindent\textbf{RQ3. Retraining Effectiveness:} finally, we investigate
whether SA can guide iterative retrainings. Is data augmentation based on high SA effective, or is adding more data simply sufficient? We augment
\train with three different sets of inputs sampled from \tradd: 1)
images with
high SA vehicle segments, 2) images with low SA vehicle segments, and 3)
randomly
chosen images with vehicle segments. These input sets are designed to evaluate the guidance provided by SA values as well as the different number of added images.
We expect the high-SA set to lead to more increased performance than the low-SA
set and include the randomly selected vehicle set as a control.
% It helps to estimate the baseline effect of simply having more training data.
% \rf{A little bit unclear why sampling from the vehicle and non-vehicle classes help here though. Try to say something more about the choice here.}
% \sy{we dropped the fourth set, mostly due to the lack of time :( }
% \rf{Ok, I changed last sentence above since it was overlapping with the one before. Please check.}
After (re-)training additional DNN models, each using augmented datasets,
we compare their performance for segmentation of the
vehicle class. It is widely believed that more diverse training data can improve
the model performance~\cite{Ma2018gf,Pei2017qy,Kim2019aa} and diverse test inputs
are important in software testing, in general~\cite{Feldt:2016if}. Since SA captures
the distance between the training data and new input, by definition,
adding additional training data with high SA values will diversify the training
data more, when compared to data with low SA values. Note that prior art
on traditional software testing indicates that diversity within the whole
set of added test data should probably be considered~\cite{Feldt:2016if}; we
leave more adaptive and complex retraining schemes for future work.\\

% \noindent\textbf{RQ4. Qualitative Assessment:} do high and low SA images share
% common features? If they do, this may be able to guide future data collection
% campaigns in specific directions, without manual labelling. For example, if
% night time driving tends to
% result in high SA images, future data collection can focus on night time
% driving to produce more such images for future trainings. We have chosen the
% top and bottom five percent of all images, in the order of their
% average IoU values across all classes, and manually inspected them to report
% any common trends.

\section{Evaluation}
\label{sec:evaluation}

This section describes the tooling we developed to apply SA to semantic
segmentation at \hm, and present the answers to RQs based on experimental
results.

\subsection{SA Analysis Pipeline}
\label{sec:implementation}

\begin{figure}[ht]
\includegraphics[width=\linewidth]{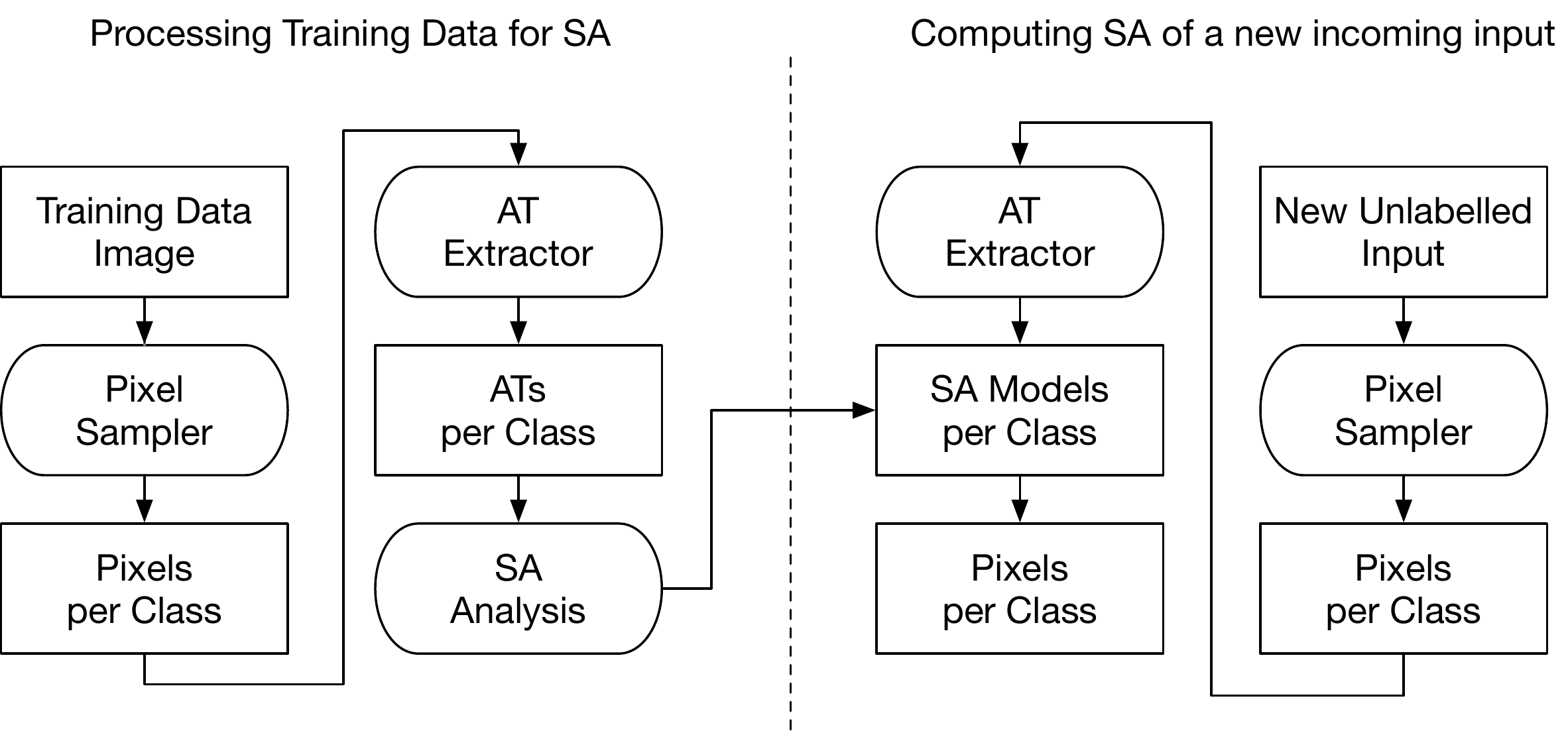}
\Description{Diagram of SA analysis process}
\caption{Overall Process of SA Analysis\label{fig:process}}
\end{figure}

\Cref{fig:process} shows the overall process of SA analysis. Given a DNN
semantic segmentation model, $M$, and the original training data, \train, we
first sample 100 pixels that belong to each class (Pixel Sampler), out of 10,000
images randomly sampled from \train. The
sampling is necessary, as each image in \train contains over 17K pixels:
processing all pixels would incur impractical computational cost, in particular for our
baseline SA technique of LSA. Once pixels
are sampled, we then extract the corresponding ATs (AT Extractor), using the
Activation Trace layer described in \Cref{sec:ats}. Since each image contains
different combinations of classes, we get 265K to 1M ATs per class. For each
class, we again sample 10K ATs to build our SA models: we compute  KDE for
LSA, or the mean vector and the covariance matrix for MDSA.

When a new unlabelled input becomes available, we first sample 1,000 pixels
per predicted class label, and extract their ATs. By running these ATs through
the SA models obtained earlier, we can compute SA values \emph{for each pixel}
in the new input. While this reflects the essence of semantic segmentation (
i.e., pixel classification), it is much more natural for us to think in terms
of images, and not pixels. Consequently, we take the mean SA of all pixels in
a specific class as the class SA of that image. The class group SA of an image
is computed as the mean of $z$-score standardised class SA values of all
classes that appear in the image. Note that this is another simplification
done for immediate industrial applicability; future work should investigate if the
distribution SA values per class and\slash or image could be leveraged for
further improvements.

\subsection{Visualisation}
\label{sec:visualisation}

Once the SA analysis is completed for a new test input, multiple \emph{views}
become available for the image: the original input image, the result of
segmentation, and an image whose pixel values are the SA values of the input
image. If the manual labelling is available, we can also place the label image,
and the difference between the model segmentation and the manual labelling. By
combining the sets of these images in the order of original video frame, we can
generate a live video visualisation of SA analysis. The side-by-side
visualisation makes it easier to compare different views, and to spot any local
high SA hotspots with their source locations in the original input.

\begin{figure}[ht]
\includegraphics[width=75mm]{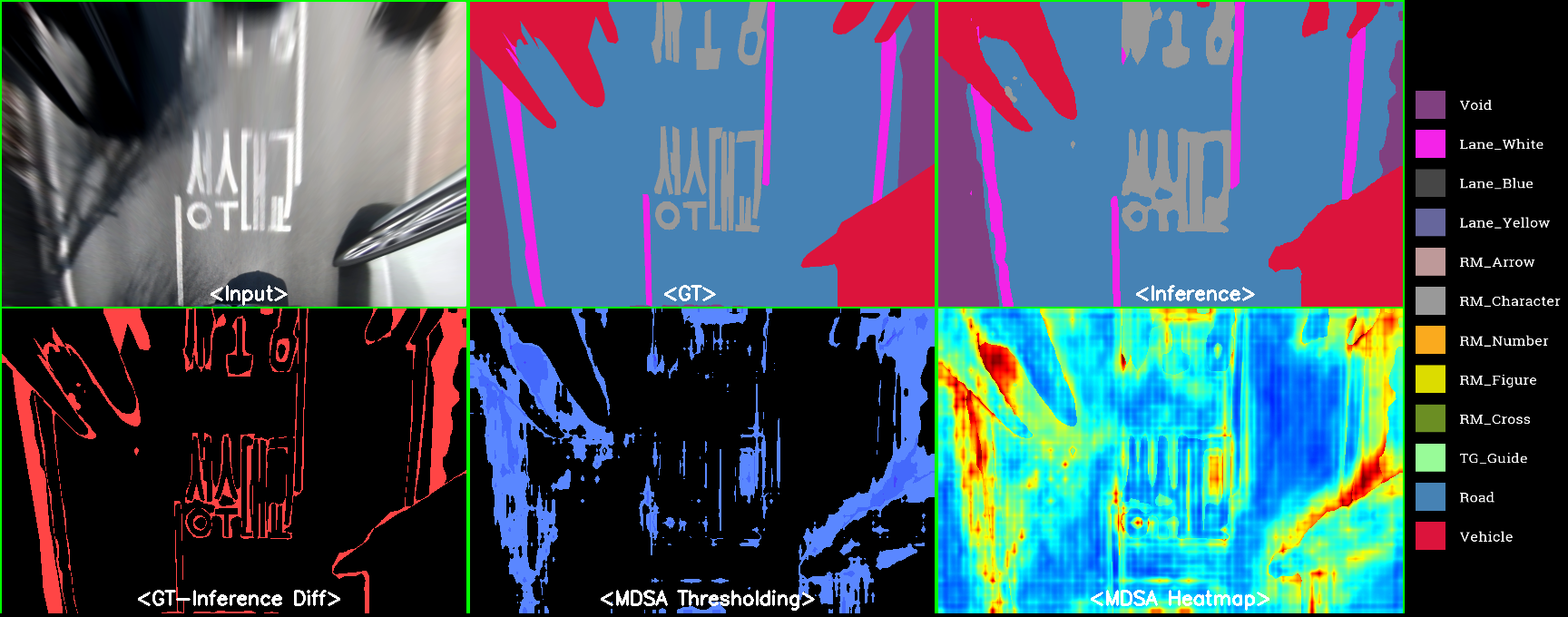}
\Description{}
\caption{Example frame of video visualisation\label{fig:visualisation}}
\end{figure}

\Cref{fig:visualisation} shows a frame from the video visualisation that
contains manual labelling information. The top row shows original input,
manual labelling, and inferred segmentation, from left to right; the bottom
row shows the diff between manual labelling and model segmentation, pixels
whose MDSA values are above a given threshold, and the MDSA heatmap. A short
video clip is available from \url{https://youtu.be/N7wKFx8pcsU}.
% \fixme{check with Hyundai whether this is doable}
% \rf{This would be asbolutely great, if possible!}\sy{They say it should be okay, but we are waiting for the final okay sign from the above...}

\subsection{Results}
\label{sec:results}

Using the analysis pipeline outlined in \Cref{sec:implementation} and the
visualisation described in \Cref{sec:visualisation}, we answer the RQs below.

\begin{figure*}[ht]
\begin{subfigure}[b]{0.45\textwidth}
\includegraphics[width=\textwidth]{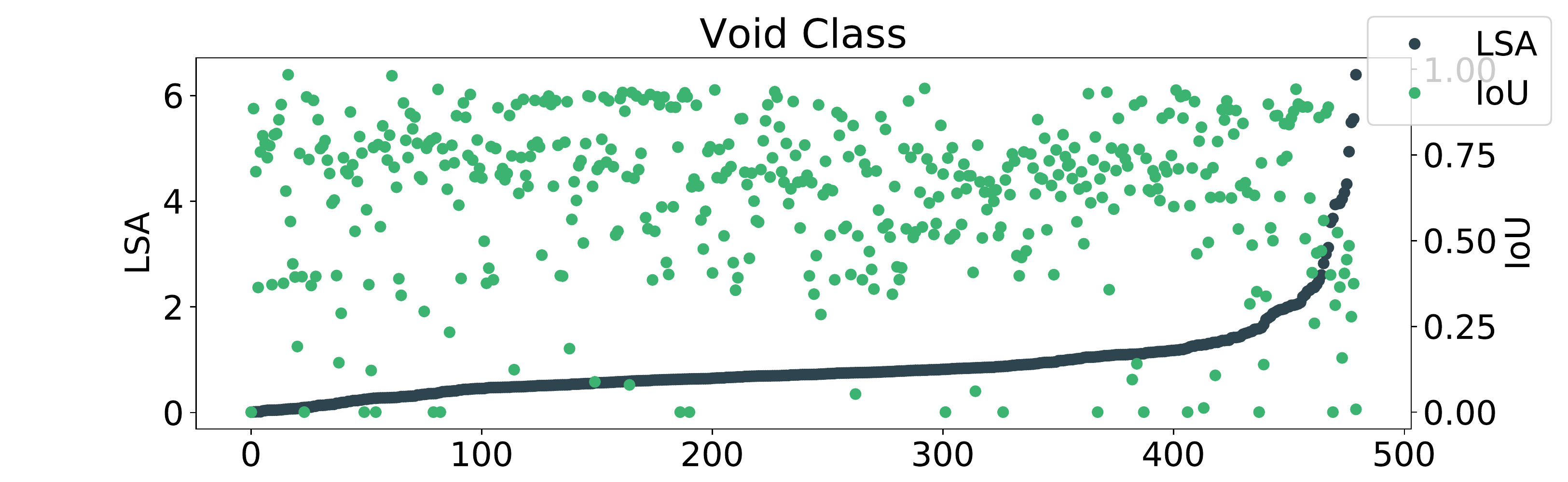}
\caption{LSA vs. IoU for Void Class\label{fig:lsa_void}}
\end{subfigure}
~
\begin{subfigure}[b]{0.45\textwidth}
\includegraphics[width=\textwidth]{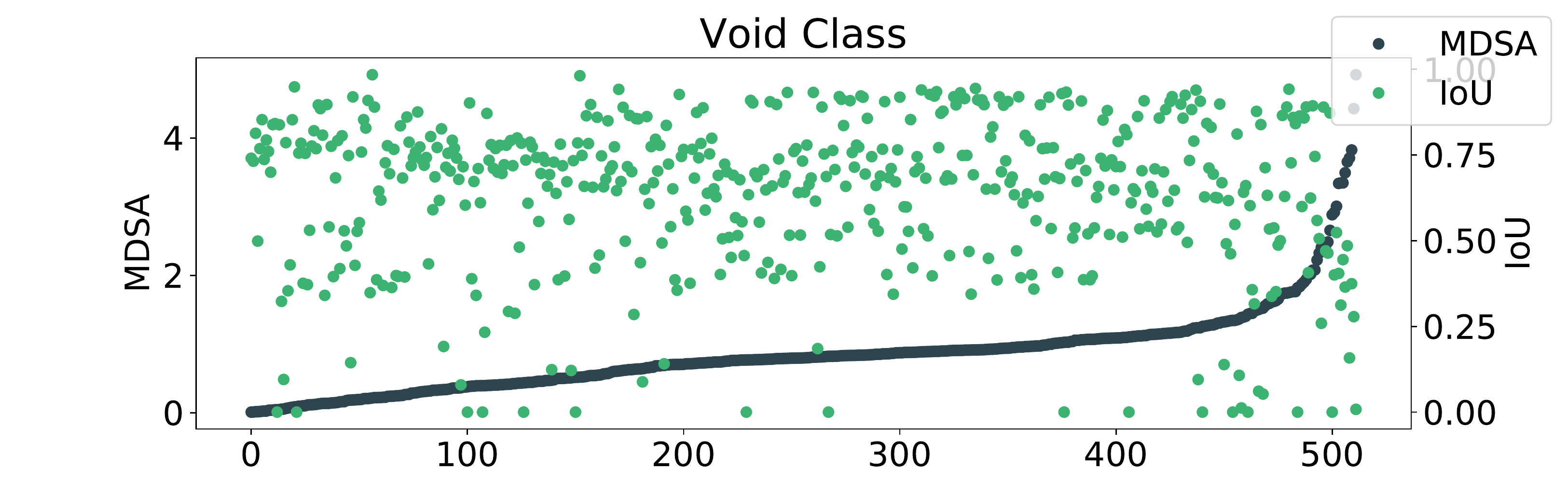}
\caption{MDSA vs. IoU for Void Class\label{fig:mdsa_void}}
\end{subfigure}

\begin{subfigure}[b]{0.45\textwidth}
\includegraphics[width=\textwidth]{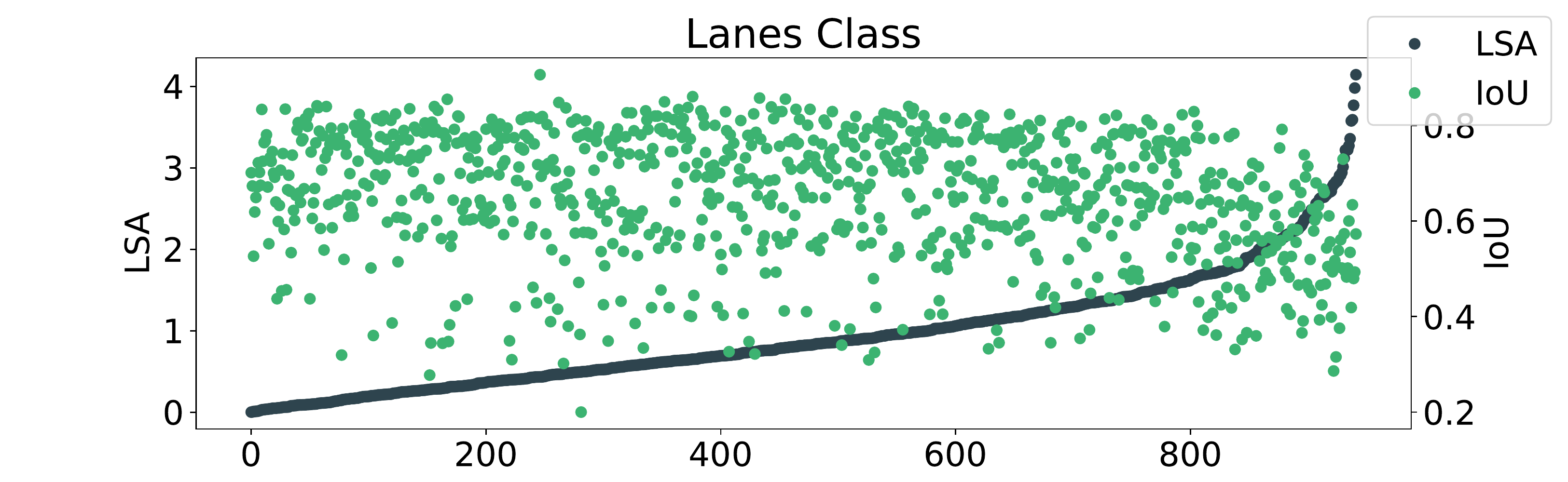}
\caption{LSA vs. IoU for Lanes Class Group\label{fig:lsa_lanes}}
\end{subfigure}
~
\begin{subfigure}[b]{0.45\textwidth}
\includegraphics[width=\textwidth]{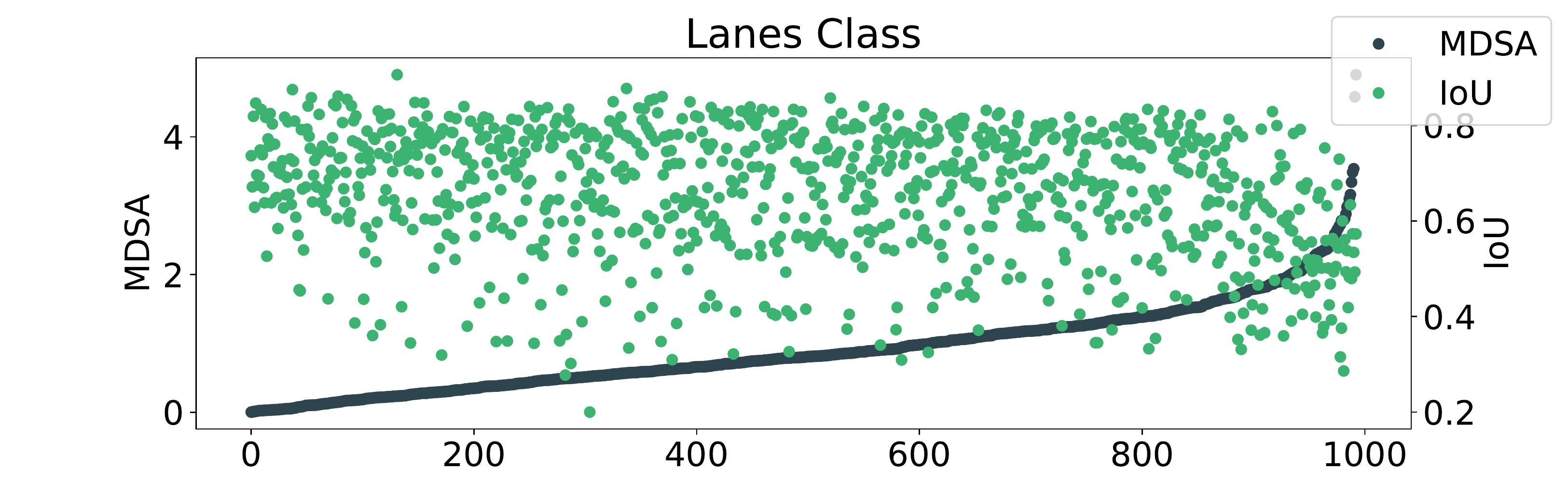}
\caption{MDSA vs. IoU for Lanes Class Group\label{fig:mdsa_lanes}}
\end{subfigure}

\begin{subfigure}[b]{0.45\textwidth}
\includegraphics[width=\textwidth]{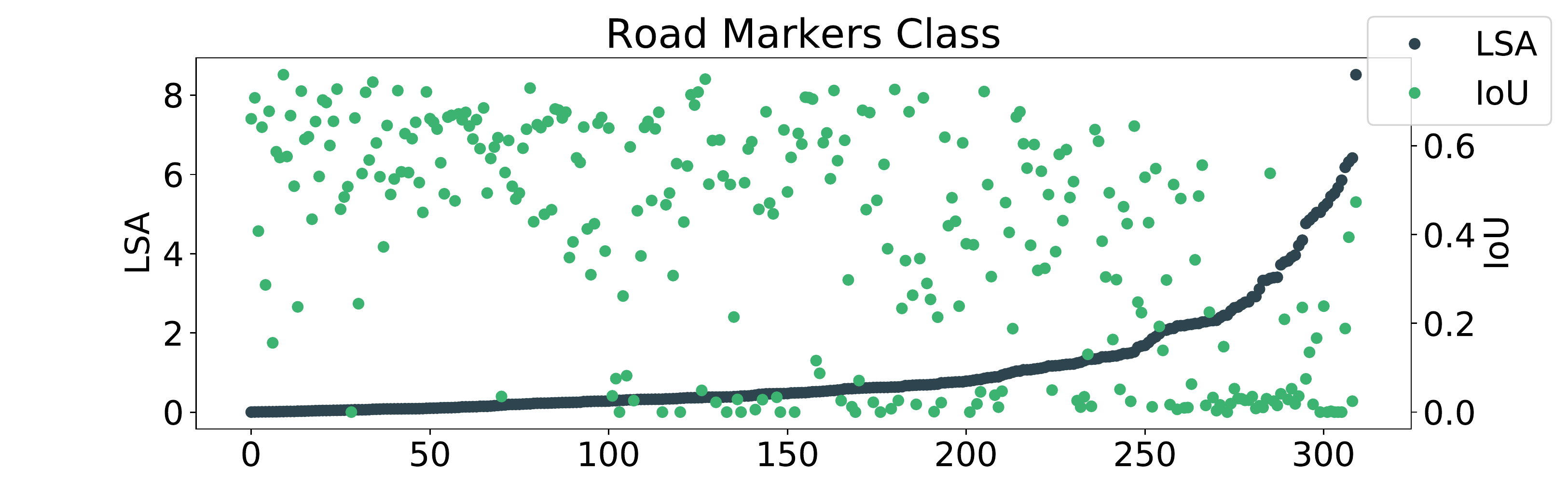}
\caption{LSA vs. IoU for Road Marker Class Group\label{fig:lsa_rm}}
\end{subfigure}
~
\begin{subfigure}[b]{0.45\textwidth}
\includegraphics[width=\textwidth]{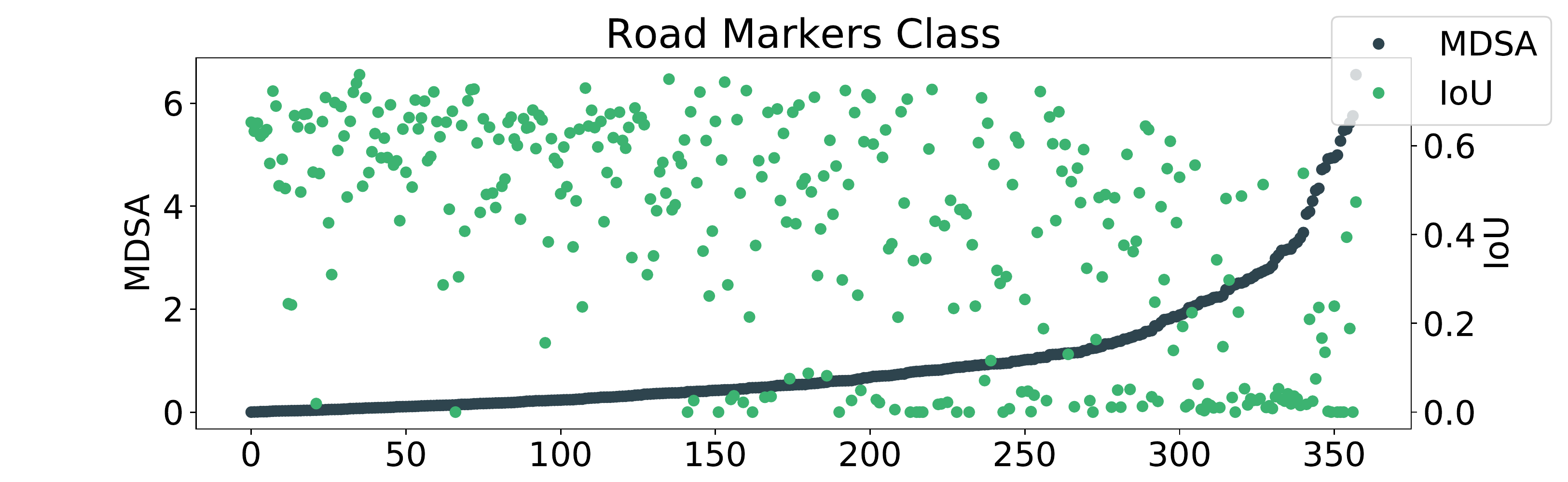}
\caption{MDSA vs. IoU for Road Marker Class Group\label{fig:mdsa_rm}}
\end{subfigure}

\begin{subfigure}[b]{0.45\textwidth}
\includegraphics[width=\textwidth]{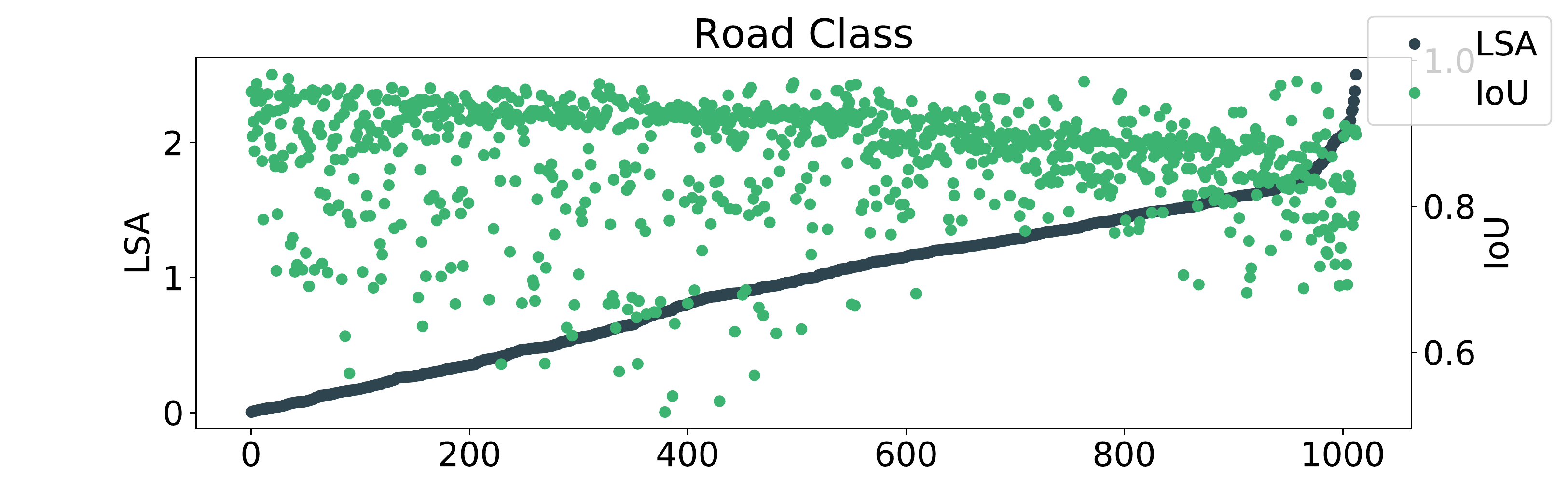}
\caption{LSA vs. IoU for Road Class\label{fig:lsa_road}}
\end{subfigure}
~
\begin{subfigure}[b]{0.45\textwidth}
\includegraphics[width=\textwidth]{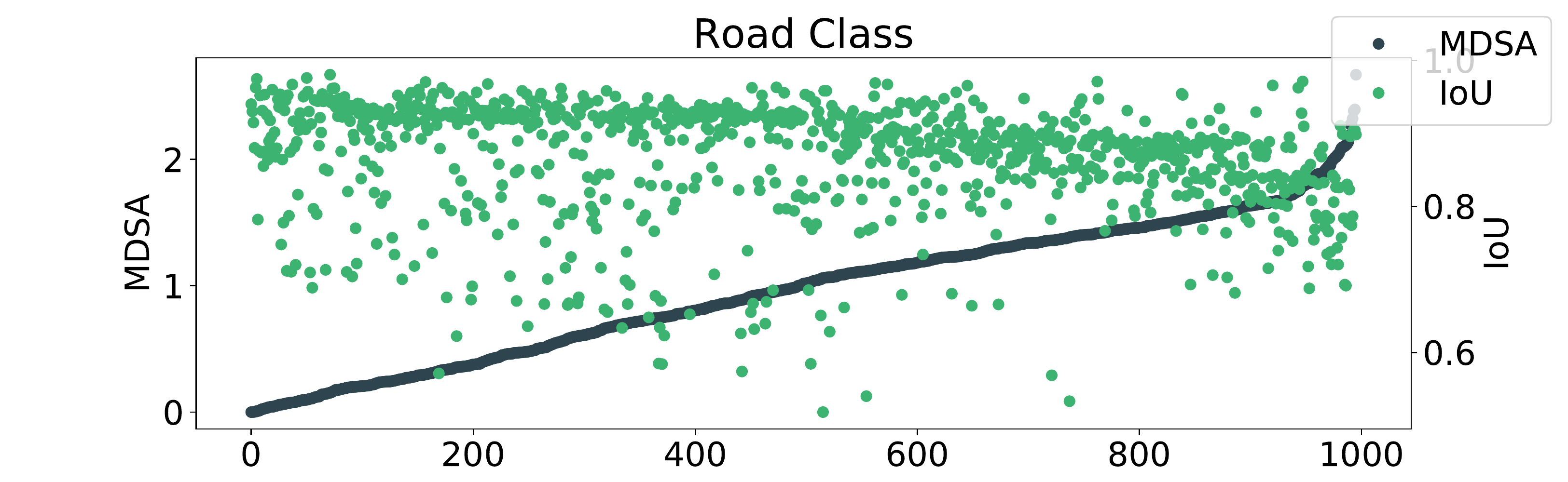}
\caption{MDSA vs. IoU for Road Class\label{fig:mdsa_road}}
\end{subfigure}

\begin{subfigure}[b]{0.45\textwidth}
\includegraphics[width=\textwidth]{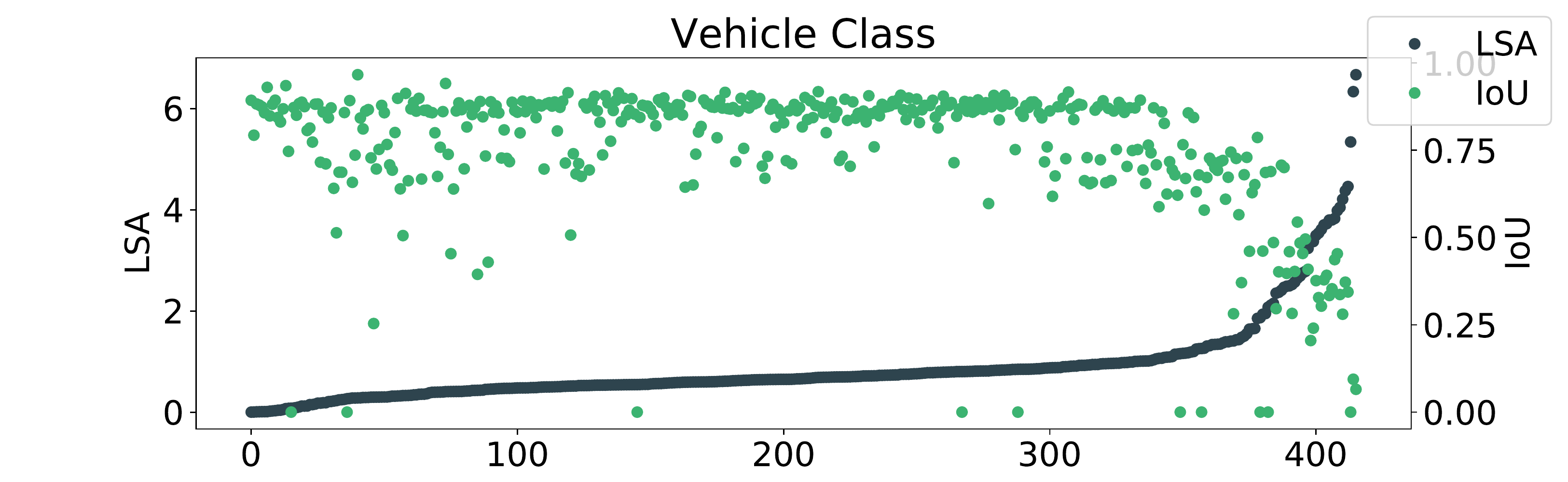}
\caption{LSA vs. IoU for Vehicle Class\label{fig:lsa_vehicles}}
\end{subfigure}
~
\begin{subfigure}[b]{0.45\textwidth}
\includegraphics[width=\textwidth]{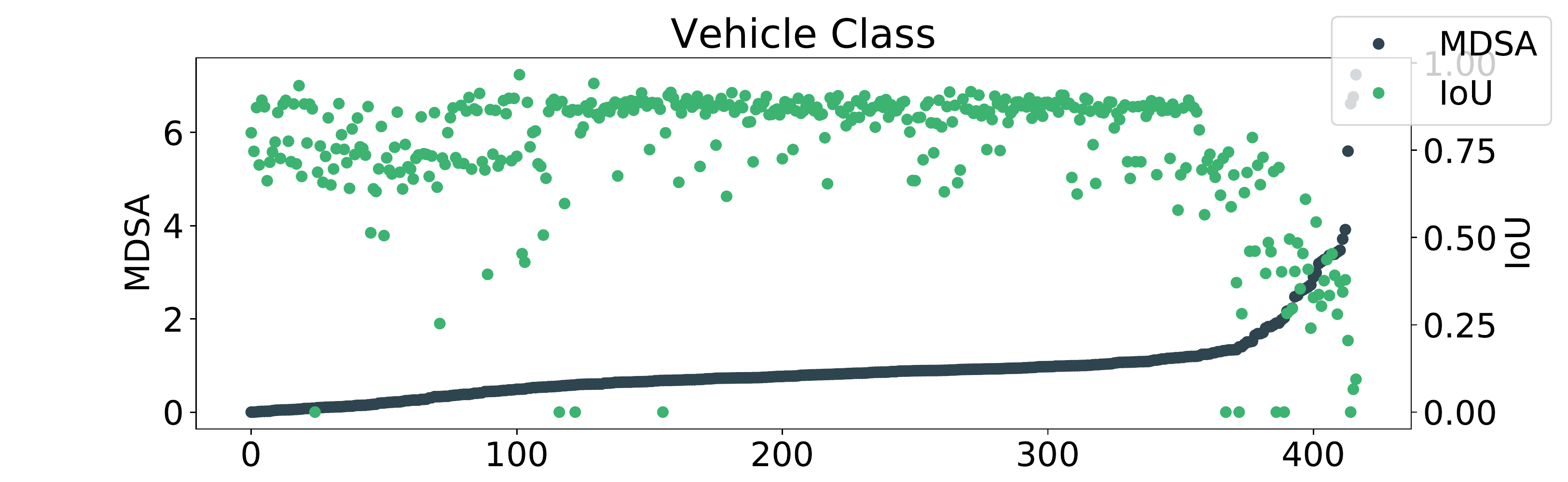}
\caption{MDSA vs. IoU for Vehicle Class\label{fig:mdsa_vehicles}}
\end{subfigure}
\Description{Plots of LSA and MDSA against IoU}
\caption{Plots of LSA and MDSA against IoU: each green circle represents the
image-wide IoU value for the given class/class group, whereas each black
circle represents the SA value of the image. We expect SA and IoU to be
negatively correlated.\label{fig:iou-vs-sa}}
\end{figure*}

\subsubsection{RQ1: Feasibility}
\label{sec:rq1}

\Cref{fig:iou-vs-sa} shows the plots of IoU against both LSA and MDSA for all
test images in \test set. For each input image in \test, one green and one black
circle are plotted: the green represents the IoU for a class or class group in
the image, whereas the black represents either the LSA or MDSA value for the
image. Based on the definition of Surprise Adequacy, we expect IoU and SA to be
negatively correlated. Basically, the higher the surprise of an input the harder
we expect the segmentation to be, thus leading to lower IoU values (remember
that the minimum IoU value would mean there is no overlap between segments, i.e.
the works possible model performance). The effect is most clearly visible in the
vehicle class (\cref{fig:lsa_vehicles} and \cref{fig:mdsa_vehicles}), but also
observed in all other classes and class groups, except the void class, which we
expected to be noisy.

\Cref{tab:correlation} shows the results of the Spearman rank correlation analysis
between SA and IoU. The second column shows the number of images that contain
the class or class group. The third and fifth columns contain $\rho_{LSA}$
and $\rho_{MDSA}$, Spearman rank correlation coefficients between the class IoU
of an image, and the class LSA/MDSA value of an image, respectively. The fourth
and the sixth columns contain the corresponding $p$-values: all correlations are
statistically significant.
Apart from the void class, which we expected to be noisy,
all other class and class groups show medium to strong negative correlation,
with correlation coefficients ranging from -0.5 to -0.715. Additionally, we
note that MDSA shows correlation values very similar to that of LSA, despite
the low computational cost. A simple profiling results suggest that MDSA is at
least two orders of magnitude faster: our LSA computation based on
\texttt{scipy}~\cite{Virtanen2020ud} can, on average, process about 249 pixels
per second, whereas the MDSA computation can process 54,980 pixels per second.
% \rf{Can we say something more concrete about the lower computational cost? At least some indication of the typical speedup for our cases here.}
% \sy{yes, think I can give some numbers here... }
For all class and class groups, the Spearman rank correlation coefficient
between MDSA and LSA values is greater than 0.98 (not shown in the table).
Based on these results, we answer RQ1 that SA analysis is successfully
applicable to semantic
segmentation: input images with high SA values tend to result in low IoU
performance. Furthermore, we also conclude that MDSA can successfully replace
LSA at a much lower cost.

\begin{table}[ht]
\caption{Spearman Rank Correlation Between SA and IoU\label{tab:correlation}}
\scalebox{0.8}{
\begin{tabular}{lrrrrr}
\toprule
Class Group  & \# of Img. & $\rho_{LSA}$ & $p$        & $\rho_{MDSA}$ & $p$        \\ \midrule
Void         & 1058 & -0.214 & 2.094e-12  & -0.224 & 1.551e-13  \\
Lanes        & 2169 & -0.5   & 9.063e-138 & -0.456 & 1.016e-111 \\
Road Markers & 1022 & -0.592 & 7.617e-98  & -0.608 & 1.662e-104 \\
Road         & 2186 & -0.715 & 0          & -0.718 & 0          \\
Vehicle      &  972 & -0.639 & 8.149e-113 & -0.556 & 7.76e-80   \\
\bottomrule
\end{tabular}}
\end{table}

\begin{figure}[ht]

  \begin{subfigure}[b]{0.4\linewidth}
  \includegraphics[width=35mm]{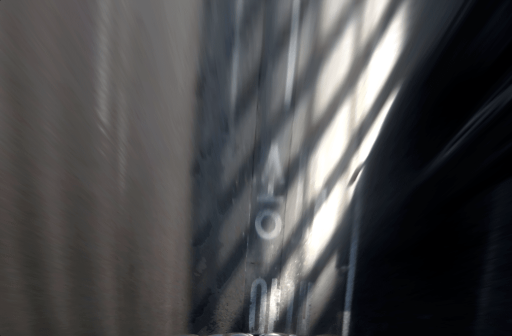}
  \caption{Shadow (High)\label{fig:shadow_input}}
  \end{subfigure}
  ~
  \begin{subfigure}[b]{0.4\linewidth}
  \includegraphics[width=35mm]{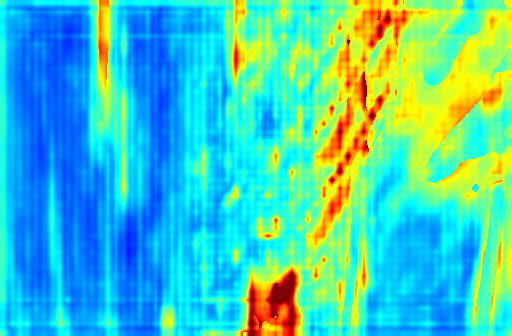}
  \caption{MDSA Heatmap\label{fig:shadow_heatmap}}
  \end{subfigure}

  \begin{subfigure}[b]{0.4\linewidth}
  \includegraphics[width=35mm]{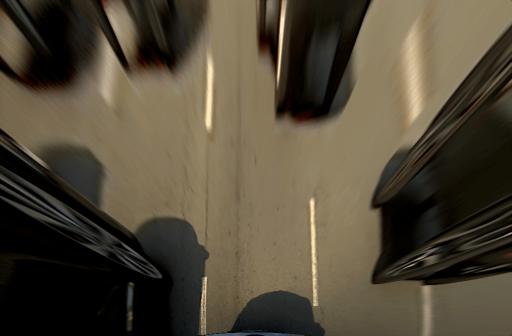}
  \caption{Cars (High)\label{fig:cars_input}}
  \end{subfigure}
  ~
  \begin{subfigure}[b]{0.4\linewidth}
  \includegraphics[width=35mm]{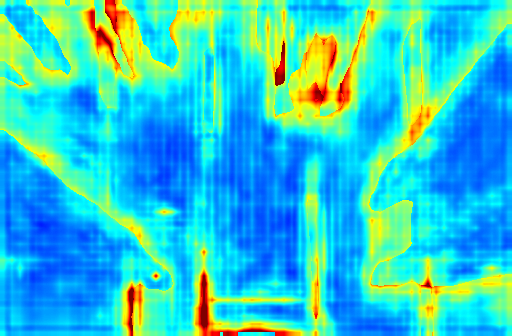}
  \caption{MDSA Heatmap\label{fig:cars_heatmap}}
  \end{subfigure}

  \begin{subfigure}[b]{0.4\linewidth}
  \includegraphics[width=35mm]{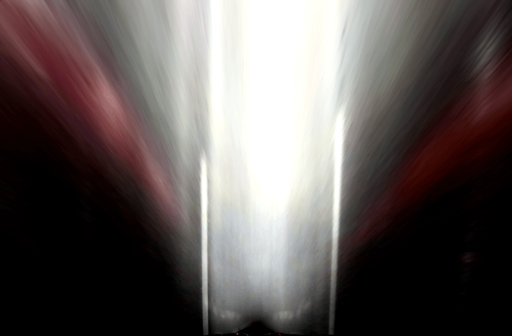}
  \caption{Night (High)\label{fig:night_input}}
  \end{subfigure}
  ~
  \begin{subfigure}[b]{0.4\linewidth}
  \includegraphics[width=35mm]{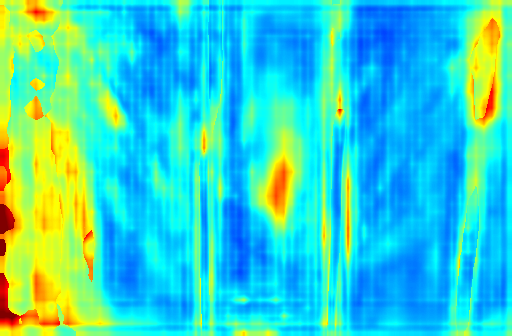}
  \caption{MDSA Heatmap\label{fig:night_heatmap}}
  \end{subfigure}

  \begin{subfigure}[b]{0.4\linewidth}
  \includegraphics[width=35mm]{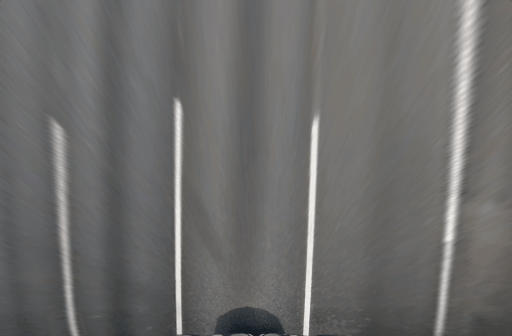}
  \caption{Road (Low)\label{fig:road_input}}
  \end{subfigure}
  ~
  \begin{subfigure}[b]{0.4\linewidth}
  \includegraphics[width=35mm]{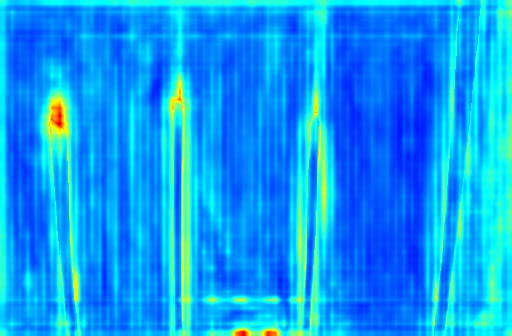}
  \caption{MDSA Heatmap\label{fig:road_heatmap}}
  \end{subfigure}
  \Description{Input images with high and low SA paired with corresponding MDSA heatmaps}
  \caption{Input images with high (\cref{fig:shadow_input}, \cref{fig:cars_input}, \cref{fig:night_input}) and low MDSA (\cref{fig:road_input}), paired with corresponding MDSA heatmaps\label{fig:rq4}}
\end{figure}

\Cref{fig:rq4} shows the representative high and low SA image examples, taken
from images with the top and bottom five percent SA values in \test. We have
identified three categories of high SA images: shadows with complex patterns,
existence of multiple vehicles, and strong headlight during night time. In the
corresponding MDSA heatmaps, the regions with high SA tend to be
where the categorical features are present. On the other hand, the low SA
images mostly contain road surfaces only, without other vehicles or shadows. Given the negative correlation between SA values
and segmentation performance, we expect engineers to be able to
plan the subsequent data collection campaigns accordingly, based on
qualitative assessments like this without manual labelling.

\subsubsection{RQ2: Cost Effectiveness}
\label{sec:rq2}

\begin{figure*}[t]
\centering
\begin{subfigure}[b]{0.5\textwidth}
\centering\includegraphics[width=\textwidth]{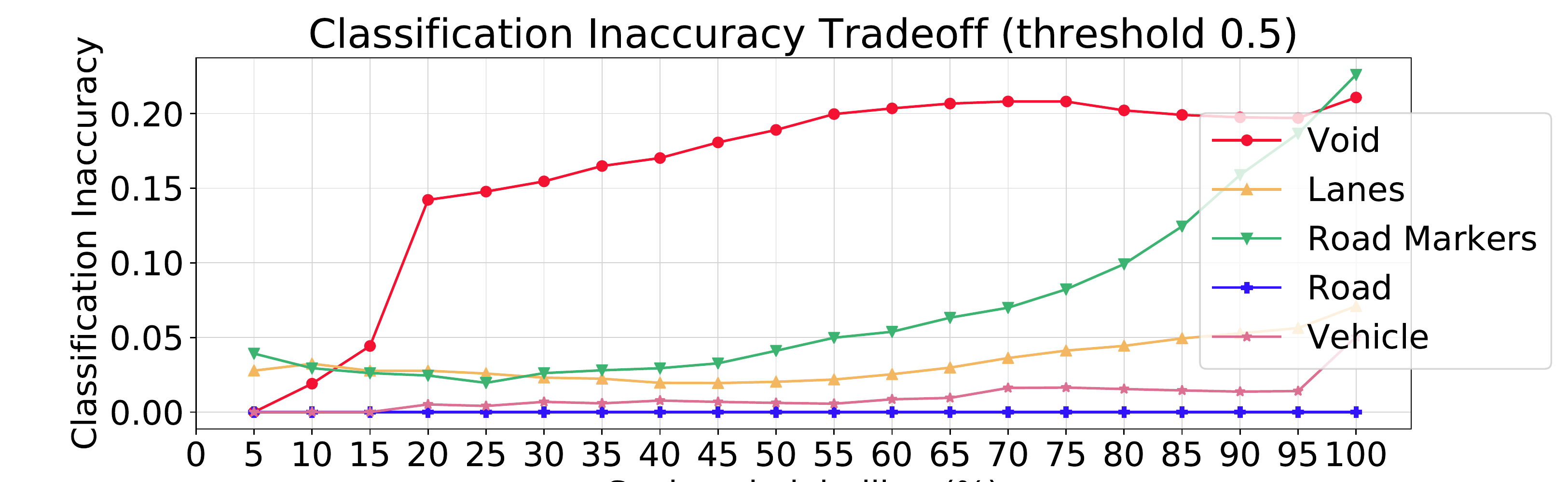}
\caption{Classification Inaccuracy with IoU Threshold 0.5\label{fig:inacc5}}
\end{subfigure}
~
\begin{subfigure}[b]{0.5\textwidth}
\centering\includegraphics[width=\textwidth]{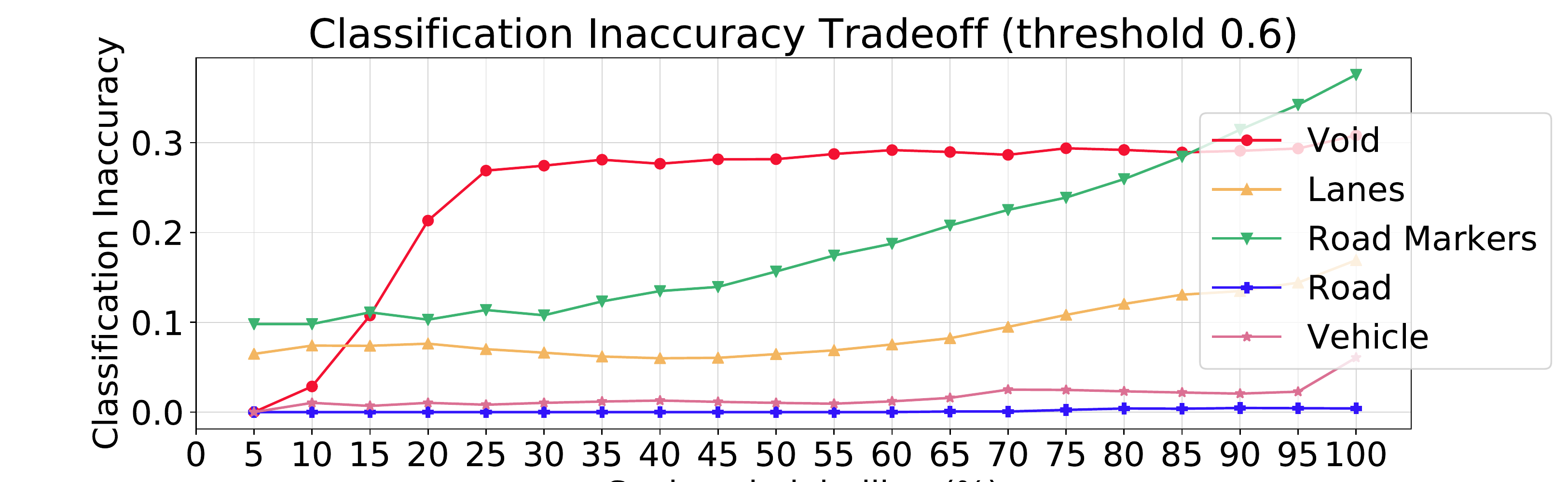}
\caption{Classification Inaccuracy with IoU Threshold 0.6\label{fig:inacc6}}
\end{subfigure}
\begin{subfigure}[b]{0.5\textwidth}
\includegraphics[width=\textwidth]{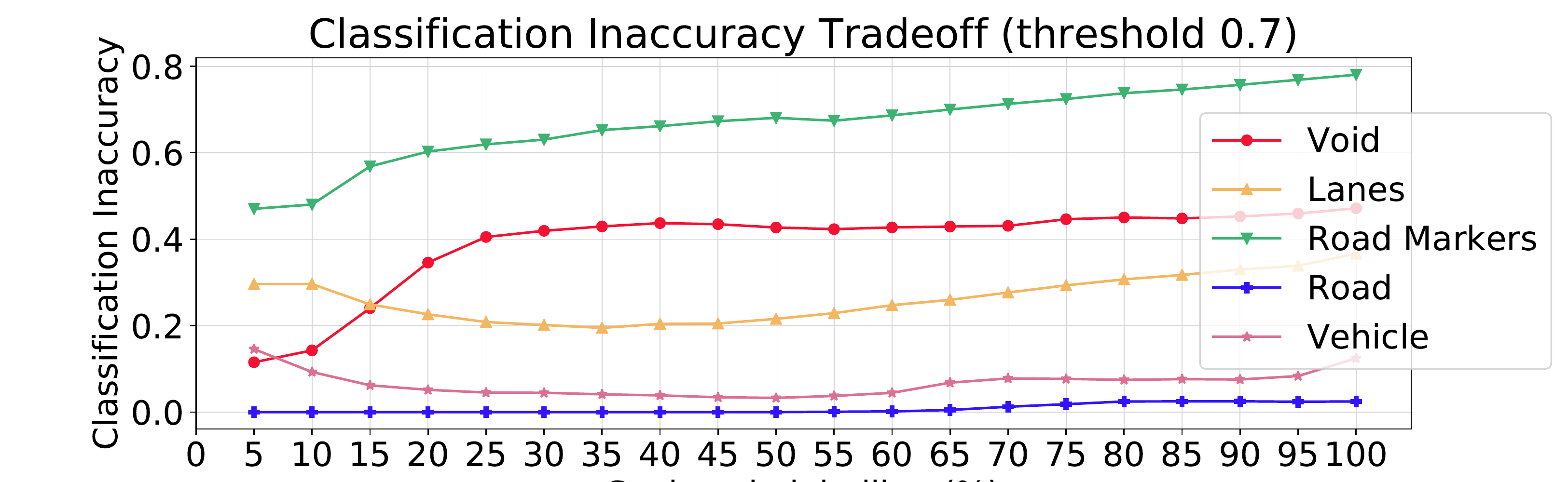}
\caption{Classification Inaccuracy with IoU Threshold 0.7\label{fig:inacc7}}
\end{subfigure}
~
\begin{subfigure}[b]{0.5\textwidth}
\includegraphics[width=\textwidth]{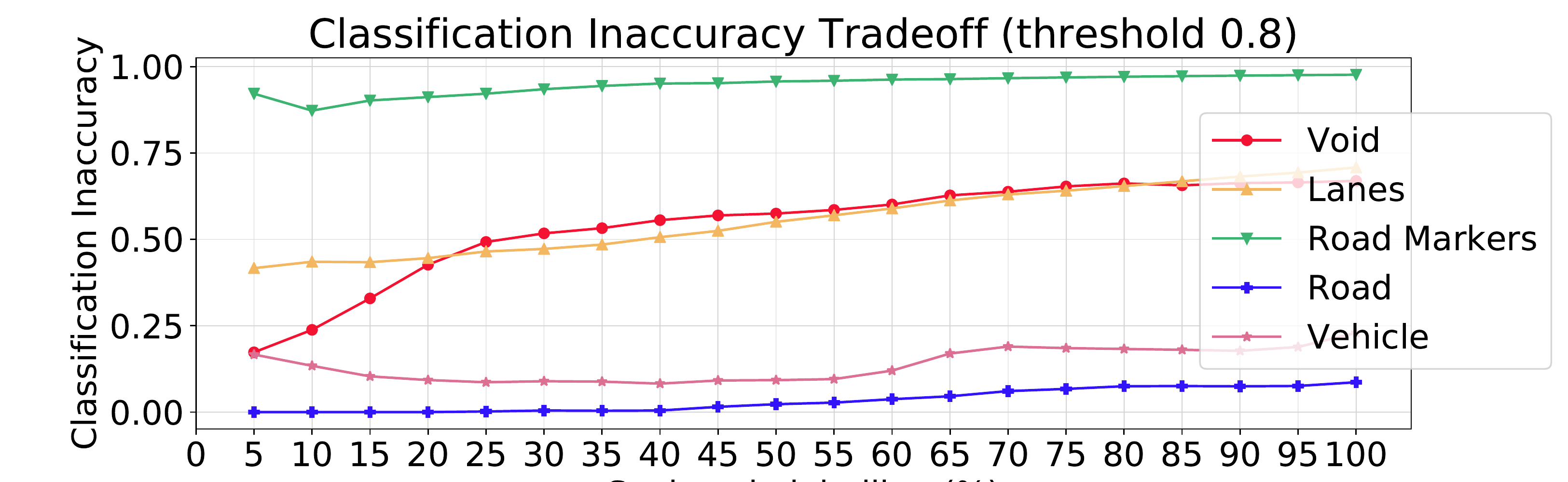}
\caption{Classification Inaccuracy with IoU Threshold 0.8\label{fig:inacc8}}
\end{subfigure}
\begin{subfigure}[b]{0.5\textwidth}
\includegraphics[width=\textwidth]{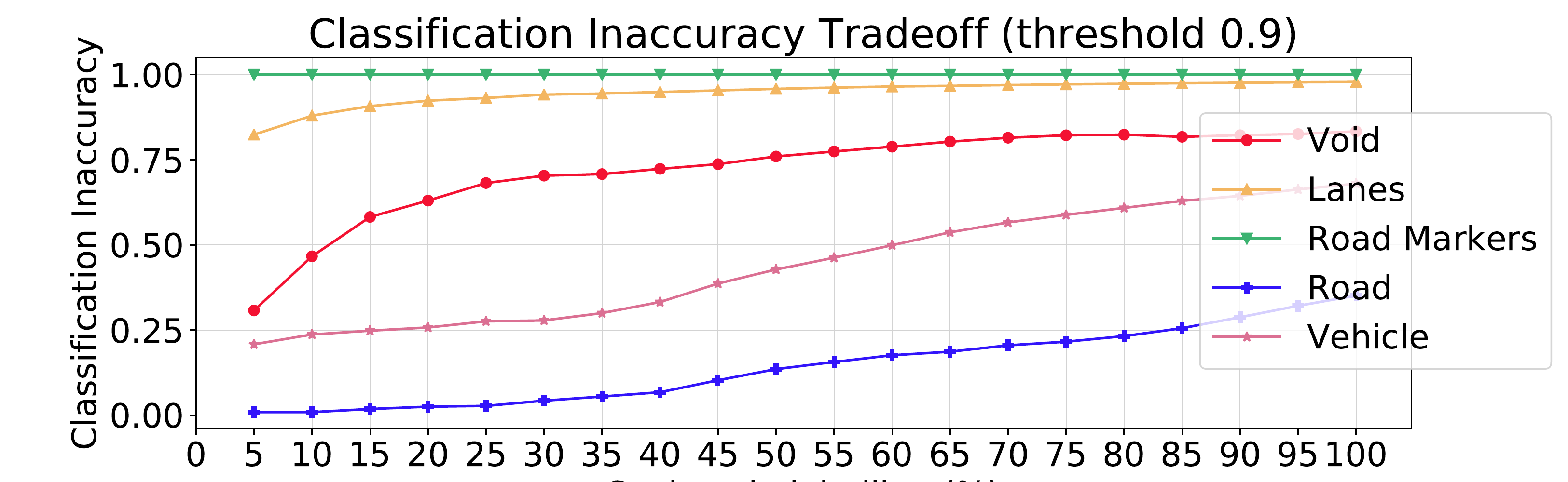}
\caption{Classification Inaccuracy with IoU Threshold 0.9\label{fig:inacc9}}
\end{subfigure}
~
\begin{subfigure}[b]{0.5\textwidth}{}
\includegraphics[width=\textwidth]{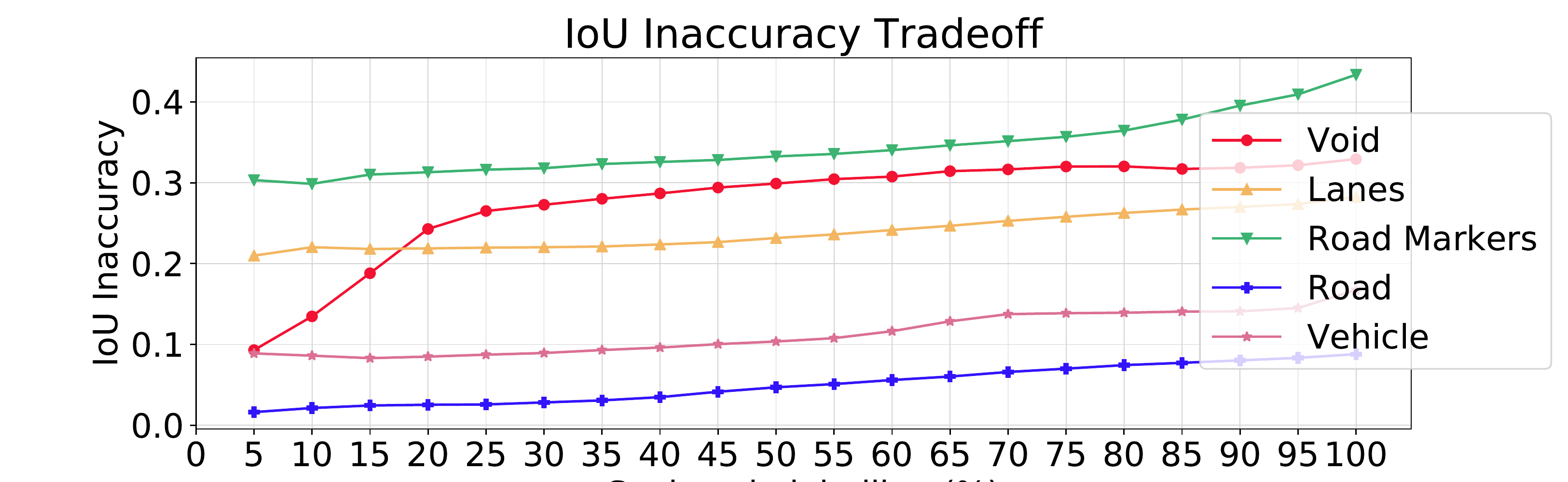}
\caption{IoU Inaccuracy\label{fig:inacciou}}
\end{subfigure}
\Description{Plots of inaccuracies against ratio of saving: $x$-axis represents the ratio of input images that will not be labelled, and $y$-axis represents the inaccuracies in the un-labelled images. The more we save (i.e., do not label), the more inaccuracy we have to accept.}
\caption{Plots of inaccuracies against ratio of saving: $x$-axis represents the ratio of input images that will not be labelled, and $y$-axis represents the inaccuracies in the un-labelled images. The more we save (i.e., do not manually label), the more inaccuracy we have to accept.\label{fig:inaccuracies}}
\end{figure*}

\Cref{fig:inaccuracies} shows the trade-off between missed inaccuracy and
labelling cost saving offered by MDSA (chosen for its superior performance
above). The $x$-axis shows the percentage of inputs we will skip to label, by
selecting them starting from the lowest MDSA values. The $y$-axis shows the
inaccuracy caused by cost saving. By not labelling these images, we are
effectively accepting the model segmentation as ground truth (i.e., IoU = 1.0),
when, in reality, they may contain errors (i.e., IoU < 1.0). As described in
\Cref{sec:rqs}, we consider images with class IoU below threshold to be
problematic, and measure the proportion of problematic input images we miss
because we do not label them, resulting in the classification inaccuracy plots
with various threshold values (\cref{fig:inacc5} to \cref{fig:inacc9}). For
example, consider the green upside down triangle ($\blacktriangledown$) at $(x,
y) = (55\%, 0.05)$ in \cref{fig:inacc5}. The data point suggests that, even when
we forego labelling 55\% of the low SA inputs, only 5\% of the skipped images
will be actually below the IoU threshold of 0.5 for the `Road Markers' class.
% \rf{Check the above example sentence, not sure I got it right ;)}
% \sy{re-written}
We also
measure the difference between the real IoU values of these images, and the
perfect IoU value of 1.0, which we assume for the sake of cost reduction,
resulting in the IoU inaccuracy plot in \cref{fig:inacciou}. The higher the
$y$-axis value is, the more inaccuracy we are forced to accept. Consequently, we
expect the inaccuracy to grow as cost saving increases. We also expect
inaccuracies to grow faster when higher IoU threshold is applied for
classification inaccuracy plots. Comparing \cref{fig:inacc5} to
\cref{fig:inacc9}, the plotted lines all move upwards, indicating higher
levels of inaccuracies and, thus, confirming the expectation.
%\rf{Clarify how fig 7f relates to the tresholds?}
%\sy{done}

The plots in \Cref{fig:inaccuracies} largely confirm our expectations yet show
surprisingly attractive trade-off. In \Cref{fig:inacciou}, the IoU inaccuracy
incurred by not labelling 40\% of the test inputs is 0.1 for the `Vehicle' class,
meaning that, on
average, the images we do not label shows IoU almost 0.9. Similarly, when we
consider images with IoU values below 0.5 threshold to be problematic, less
than 5\% of images we skip to label will be actually problematic with vehicle,
lanes, and road maker class and class groups, when we save up to 50\% of
labelling cost, see \cref{fig:inacc5}. Note that, while we are studying the trade-off
with the hindsight of having the results of manual labelling, the actual
decision to save labelling cost can be made solely based on SA values without
the labelling. The amount of saving can be guided either by a study like this,
or by a pilot study about the trade-off that is unique to the DNN model being
developed. Based on the observed trade-off, we answer RQ2 that SA can be
successfully used to reduce the cost of manual labelling.

\subsubsection{RQ3. Retraining Effectiveness}

To answer RQ3, we compare the impact of adding high and low SA images to the
base training data. In addition to \train and \test, From a large number of
images from a separate and independent data collection campaign has been made
available for RQ3. We divided this pool of 70,532 images into \tradd and
\teadd. The set \teadd will be used to evaluate our retraining; the set \tradd has been further divided into multiple subsets as follows.
\begin{itemize}
\item \tradd[\textbf{B}ase] set: we randomly chose 16,750 images out of \tradd
to use as the base training set for retraining.
\item \tradd[SA Level, SIZE] sets: we chose images that have more than 1,000 pixels
labelled as vehicle class out of \tradd $\setminus$ \tradd[Base], resulting in
24,898 images.
We then compute SA for these 24,898 images,
sample nine subsets based on parameters Size and SA Level, and add them to
\tradd[Base]. The resulting datasets are denoted by \tradd[SA Level, Size].
Size can be either `\textbf{L}arge', `\textbf{M}edium', and `\textbf{S}mall':
a `Large' subset
contains 5,025 images (30\% of 16,750), a `Medium' 3,375 images (20\% of
16,750), and a `Small' 1,675 images (10\% of 16,750). SA Level can be either
`\textbf{H}igh', `\textbf{L}ow', or `\textbf{R}andom': a `High' subset is sampled from the top 30\%
images sorted in the descending order of SA values, a `Low' subset from the
bottom 30\%, and a `Random' subset completely randomly. Finally, we add
each of these nine subsets to \tradd[\textbf{B}ase], resulting in nine datasets.
\end{itemize}
% \rf{Good!}

% \rf{Check how this descriptions aligns with the previous one about the experimental
% setup; a bit confusing to me why both are needed. Can this be moved the the earlier section?
% Also what happened to the set with no-vehicle images mentioned earlier? I couldn't fix this since not clear to me how things align...}
In total, we generate ten new training datasets, and perform 10 (re)trainings,
using the same set of hyperparameters used in RQ1 and RQ2. Subsequently, we
compare the average vehicle class IoU, obtained by each of the trained models,
using images in \teadd.
% \fixme{check with Hyundai whether this is actually the case}

\Cref{tab:retraining_iou_body} and \Cref{tab:retraining_iou_wheel} show the vehicle body and wheel class IoUs from the nine retrainings with additional
data, along with the corresponding IoU from the original model trained
using \tradd[\textbf{B}ase]. Row-wise maximum values are typeset in bold,
while column-wise maximum values are underlined. In all cases, augmented
datasets resulted in higher IoU than the result from \tradd[\textbf{B}ase].
However, the comparison between the augmented results reveals an interesting
interplay between the size of the added datasets and their SA levels.

Column-wise, larger datasets tend to produce higher IoU values: this pattern
is observed for high SA and random datasets in both
\Cref{tab:retraining_iou_body} and \Cref{tab:retraining_iou_wheel}. Row-wise,
adding high SA images is more effective when we are adding a smaller number of
images (e.g., compare \tradd[H, S] with \tradd[R, S]). However, when more
images become available, the SA level has less impact: this can be seen from
the comparison of \tradd[H, L] and \tradd[R, L]. While results are limited due
to the practical constraint on the number of retrainings we could perform,
we cautiously suggest that high SA images are more impactful on retraining
when only relatively fewer new images can be added. However, with a sufficient
number of new images, the diversity within the new images appears to take over.
This is in line with general results on diversity-driven
testing~\cite{Feldt:2016if}, where the diversity is most important early on
during test selection.
Thus, this calls for a new augmentation technique that considers both the diversity
with respect to the training dataset (captured by SA) and the diversity within
the data being augmented (currently not being captured as a metric, but set-based
metrics \cite{Feldt:2016if} might help),
as well as a larger scale empirical evaluation. We answer
RQ2 that, while data augmentation based on high SA is effective for small
sized augmentations, the effectiveness of data augmentation shows complex
interplay between the size and the SA, calling for further investigation.

\begin{table}
\caption{IoU for vehicle body class after retraining with various additional training datasets: {\normalfont \textbf{bold} and \underline{underlined} represent row and column maximum.}
\label{tab:retraining_iou_body}}
\scalebox{0.85}{
\begin{tabular}{lrlrlr}
\toprule
Dataset      & IoU    & Dataset      & IoU    & Dataset      & IoU    \\ \midrule
\tradd[B]    & 0.3831             & \tradd[B]    & 0.3831             & \tradd[B]    & 0.3831 \\
\tradd[H, S] & \textbf{0.4329}             & \tradd[L, S] & 0.4305             & \tradd[R, S] & 0.4246 \\
\tradd[H, M] & 0.4253             & \tradd[L, M] & \underline{\textbf{0.4397}} & \tradd[R, M] & 0.4269 \\
\tradd[H, L] & \underline{0.4392} & \tradd[L, L] & 0.4359             & \tradd[R, L] & \underline{\textbf{0.4417}} \\
\bottomrule
\end{tabular}
}
\end{table}

\begin{table}
\caption{IoU for vehicle wheel class after retraining with various additional training datasets: {\normalfont \textbf{bold} and \underline{underlined} represent row and column maximum.}
\label{tab:retraining_iou_wheel}}
\scalebox{0.85}{
\begin{tabular}{lrlrlr}
\toprule
Dataset      & IoU    & Dataset      & IoU    & Dataset      & IoU    \\ \midrule
\tradd[B]    & 0.3640             & \tradd[B]    & 0.3640                      & \tradd[B]    & 0.3640          \\
\tradd[H, S] & \textbf{0.4175}    & \tradd[L, S] & 0.4158                      & \tradd[R, S] & 0.4132          \\
\tradd[H, M] & 0.4125             & \tradd[L, M] & \underline{\textbf{0.4242}} & \tradd[R, M] & 0.4160          \\
\tradd[H, L] & \underline{0.4250} & \tradd[L, L] & 0.4217                      & \tradd[R, L] & \underline{\textbf{0.4302}} \\
\bottomrule
\end{tabular}
}
\end{table}

\section{Threats to Validity}
\label{sec:threats}

Threats to internal validity include the correctness of the studied models as
well as the SA analysis pipeline. To mitigate threats, all the model we study
has been internally trained and validated at \hm by multiple domain experts;
the SA analysis pipeline follows the same approach that has been publicly
replicated and reproduced~\cite{Kim2019aa}. Threats to external validity
concern any issues that may restrict the degree to which the results
generalise. While the observed results are all specific to the models and the
data we studied, the SA analysis has been shown to work with data from other
domains~\cite{Kim2019aa}, and also worked as expected when applied to separate,
independent dataset for RQ3. Our analysis pipeline applies aggressive sampling
to reduce computational cost and, therefore, to increase the practical
applicability. If more powerful computational resources are available, we
expect processing more data will improve the accuracy of SA analysis. The
retraining experiment for RQ3 was constrained by the available computational
resources. We will conduct a larger scale empirical evaluation with novel
augmentation techniques as future work. Finally,
threats to construct validity concern whether we are measuring what we claim
to measure. To mitigate this concern, we primarily use intersection over union,
which is a well understood, standard evaluation metric for semantic
segmentation task.

\section{Related Work}
\label{sec:relatedwork}

There are many existing work on testing of DNN models.
DeepXplore~\cite{Pei2017qy} and DeepGauge~\cite{Ma2018ny} introduced multiple
coverage criteria that measure the diversity of input sets: using a more diverse
set of inputs is more likely to reveal misbehaviours of the DNN under test.
DeepTest~\cite{Tian2018aa} focuses on improving Neuron Coverage, one of the
coverage criteria proposed in DeepXplore, by applying systematic perturbations
to existing input images. Surprise Adequacy~\cite{Kim2019aa} proposed test
adequacy based on the distance between a single new input and the set of inputs
in the training data, thereby enabling the comparison between individual inputs.
Since labelling cost depends on the number of input images, we use SA to make
the labelling decision for individual input images.

Another body of existing work on DNN testing focuses on improving the model
performance. MODE~\cite{Ma2018gf} aims to \emph{debug} model misbehaviour by
selecting inputs that are relevant to misbehaviour for retraining.
Apricot~\cite{Zhang2019zj}, on the other hand, first trains multiple DNN models
using reduced datasets (called rDLMs), and directly manipulates the neural
weights of the misbehaving DNN towards the average of the weights of correctly
behaving rDLMs and away from the average of those of misbehaving rDLMs. Our
approach in this paper is close to that of MODE, in that we go through
retraining instead of manipulating the model weights directly. However, our
focus is on choosing inputs to use with a focus on the cost of selecting them,
which, in turn, depends on the labelling cost. As far as we know, this is the
first industry case study that looks at the trade-off between manual labelling
cost and accuracy of DNN evaluation/retraining. Similarly, work on
active learning for reducing labelling effort~\cite{Beluch2018pt,Gal2017rq}
are typically lab experiments or done on publicly available datasets rather
than in industrial practice.

\section{Conclusion and Future Work}
\label{sec:conclusion}

We propose a technique to reduce manual labelling cost during the development
of a DNN based semantic segmentation module for autonomous driving in the
automotive industry.
Semantic segmentation has a high cost of labelling, as the act of manual
labelling involves high precision pixel classification with high resolution
input images. We exploit the negative correlation between Surprise Adequacy (SA)
and model performance to decide for which images we can skip labelling. We also use
SA to guide the selection of newly collected inputs to be added to the base
training dataset, so that the model performance can be effectively improved.
The proposed technique is evaluated in an industry case study involving a real
world semantic segmentation DNN model and actual road data, both trained and
collected by \hm. The result shows that 30 to 50\% of manual labelling cost can
be saved with negligible impact on evaluation accuracy, and that SA can effectively
guide input selection for retraining without incurring high labelling cost.

The proposed labelling cost reduction technique used a fixed threshold (i.e.,
top x\% of low SA inputs). Future work will consider more intelligent and
adaptive decision mechanism to decide whether to manually label a new incoming
input or not. This mechanism can rely on SA as well as other features of the
input image to realise even more effective cost reduction while minimising
evaluation inaccuracy. Future work should also explore other application domains
and tasks since our base technique is fully general and not specific to
semantic segmentation.

\section*{Acknowledgement}

Jinhan Kim and Shin Yoo have been supported by Hyundai Motors, Next-Generation
Information
Computing Development Program through the National Research Foundation of Korea
(NRF) funded by the Ministry of Science, ICT (2017M3C4A7068179), Engineering
Research Center Program through the National Research Foundation of Korea (NRF)
funded by the Korean Government MSIT (NRF-2018R1A5A1059921), and Institute of
Information \& communications Technology Planning \& Evaluation (IITP) grant
funded by the Korea government(MSIT) (No.2018-0-00769, Neuromorphic Computing
Software Platform for Artificial Intelligence Systems). Robert Feldt has been
supported by the Swedish Scientific Council (No.2015-04913, Basing Software
Testing on Information Theory).

\bibliographystyle{ACM-Reference-Format}
\bibliography{newref}

\end{document}